\documentclass[10pt,twocolumn,letterpaper]{article}

\usepackage{cvpr}              

\definecolor{cvprblue}{rgb}{0.21,0.49,0.74}
\usepackage[pagebackref,breaklinks,colorlinks,allcolors=cvprblue]{hyperref}
\usepackage{times}  
\usepackage{helvet}  
\usepackage{courier}  
\usepackage[hyphens]{url}  
\usepackage{graphicx} 
\usepackage{natbib}  
\usepackage{caption} 
\usepackage{algorithm}
\usepackage{algorithmic}
\usepackage[accsupp]{axessibility}
\usepackage{xcolor}
\usepackage[table]{xcolor} 
\usepackage{makecell}
\usepackage{amsmath}
\usepackage{amssymb}
\usepackage{subcaption}
\usepackage[most]{tcolorbox}
\usepackage{booktabs}
\usepackage{multirow}
\usepackage{colortbl}
\usepackage{pifont}

\usepackage{newfloat}
\usepackage{listings}
\definecolor{mygray}{gray}{.9}
\definecolor{mygreen}{RGB}{93,173,85}
\definecolor{mywarning}{RGB}{233,144,61}

\definecolor{DarkBlue}{RGB}{64,101,149}
\definecolor{azure}{rgb}{0.0, 0.5, 1.0}
\definecolor{gray}{rgb}{0.3, 0.3, 0.3}
\definecolor{DarkGreen}{RGB}{42,110,63}
\definecolor{Green}{rgb}{0.0, 0.5, 0.0}       
\definecolor{RedOrange}{rgb}{1.0, 0.27, 0.0}  

\newcommand{\myred}[1]{$_{\color{RedOrange}\uparrow #1}$}       

\newcolumntype{x}[1]{>{\centering\arraybackslash}p{#1pt}}
\newcolumntype{I}{!{\vrule width 1pt}} 
\tcbuselibrary{theorems}
\definecolor{lightgray}{gray}{.9}
\definecolor{deepgray}{gray}{.8}
\tcbset{highlight math/.append style={left=0mm,right=0mm,top=0mm,bottom=0mm, colframe=white}}
\definecolor{keywordcolor}{RGB}{178,34,34} 
\newcolumntype{x}[1]{>{\centering\arraybackslash}p{#1pt}}
\newcolumntype{I}{!{\vrule width 1pt}} 
\tcbuselibrary{theorems}
\definecolor{lightgray}{gray}{.9}
\definecolor{deepgray}{gray}{.8}
\tcbset{highlight math/.append style={left=0mm,right=0mm,top=0mm,bottom=0mm, colframe=white}}
\definecolor{keywordcolor}{RGB}{178,34,34} 
\makeatletter
\newcommand{\thickhline}{%
    \noalign {\ifnum 0=`}\fi \hrule height 1pt
    \futurelet \reserved@a \@xhline
}
\newtcolorbox{casebox}[1]{
  enhanced,
  colback=black!2,
  colframe=black!60,
  boxrule=0.6pt,
  arc=2mm,
  outer arc=2mm,
  top=10mm,
  bottom=4mm,
  left=8mm,
  right=8mm,
  fontupper=\ttfamily\small,
  width=\textwidth,
  title={#1},
  overlay unbroken={
    \path[fill=black!75, draw=black!75]
      (frame.north west) rectangle ([yshift=-18pt]frame.north east);
    \node[anchor=west, font=\bfseries\ttfamily\large, text=white]
      at ([xshift=10pt, yshift=-9pt]frame.north west)
      {\tcbtitle};
  }
}

\newtcolorbox{caseboxpurple}[1]{%
  enhanced,
  colback=violet!5,          
  colframe=violet!50!black,  
  boxrule=0.6pt,
  arc=2mm,
  outer arc=2mm,
  top=12mm,
  bottom=4mm,
  left=8mm,
  right=8mm,
  fontupper=\ttfamily\small, 
  width=\textwidth,
  title={#1},
  overlay unbroken={
    \path[fill=violet!70!black, draw=violet!70!black]
      (frame.north west) rectangle ([yshift=-18pt]frame.north east);
    \node[anchor=west, font=\bfseries\ttfamily\large, text=white]
      at ([xshift=10pt, yshift=-9pt]frame.north west)
      {\tcbtitle};
  }
}

\newtcbox{\predwrong}{
  on line,
  colback=red!10,
  colframe=red!60!black,
  boxrule=0.4pt,
  arc=1mm,
  left=1pt,
  right=1pt,
  top=1pt,
  bottom=1pt
}

\newtcbox{\predcorrect}{
  on line,
  colback=green!10,
  colframe=green!60!black,
  boxrule=0.4pt,
  arc=1mm,
  left=1pt,
  right=1pt,
  top=1pt,
  bottom=1pt
}

\newcommand{\cmark}{\ding{51}} 
\newcommand{\xmark}{\ding{55}} 
\makeatother

\def\paperID{12631} 
\def\confName{CVPR}
\def\confYear{2026}

\title{\texttt{Mario}: Multimodal Graph Reasoning with Large Language Models}

\author{
Yuanfu Sun$^{1,2}$\footnotemark[1], Kang Li$^{3}$\footnotemark[1], Pengkang Guo$^{4}$, Jiajin Liu$^{1,2}$, Qiaoyu Tan$^{1}$\footnotemark[2]\\
$^{1}$New York University Shanghai, 
$^{2}$New York University, 
$^{3}$Tsinghua University, 
$^{4}$EPFL\\
{\small \texttt{\{yuanfu.sun, qiaoyu.tan\}@nyu.edu, lik24@mails.tsinghua.edu.cn}}
}

\begin{document}
\maketitle
\renewcommand{\thefootnote}{\fnsymbol{footnote}}
\footnotetext[1]{Equal contribution} \footnotetext[2]{Corresponding author}

\begin{abstract}
Recent advances in large language models (LLMs) have opened new avenues for multimodal reasoning. Yet, most existing methods still rely on pretrained vision–language models (VLMs) to encode image–text pairs in isolation, ignoring the relational structure that real-world multimodal data naturally form. This motivates reasoning on multimodal graphs (MMGs), where each node has textual and visual attributes and edges provide structural cues. Enabling LLM-based reasoning on such heterogeneous multimodal signals while preserving graph topology introduces two key challenges: resolving weak cross-modal consistency and handling heterogeneous modality preference. To address this, we propose Mario, a unified framework that simultaneously resolves the two above challenges and enables effective LLM-based reasoning over MMGs. Mario consists of two innovative stages. Firstly, a graph-conditioned VLM design that jointly refines textual and visual features through fine-grained cross-modal contrastive learning guided by graph topology. Secondly, a modality-adaptive graph instruction tuning mechanism that organizes aligned multimodal features into graph-aware instruction views and employs a learnable router to surface, for each node and its neighborhood, the most informative modality configuration to the LLM. 
Extensive experiments across diverse MMG benchmarks demonstrate that Mario consistently outperforms state-of-the-art graph models in both supervised and zero-shot scenarios for node classification and link prediction. The code will be made available at \href{https://github.com/sunyuanfu/Mario}{Mario}.

\end{abstract}

\section{Introduction}
\noindent
Large language models (LLMs) \cite{liu2024deepseek,guo2025deepseek,yang2024qwen2} have evolved from pure text processors into general-purpose multimodal reasoners that can follow instructions, ground entities, and integrate visual signals \cite{peng2023kosmos}. Yet most current pipelines \cite{liu2025graph,pan2022contrastive,yoon2023multimodal,wei2019mmgcn} continue to assume and operate on a simplified input form, in which multimodal data are handled as isolated image–text pairs. This view sits uneasily with how multimodal data actually appear in the wild. Rather than isolated image–text pairs, these entities are interlinked and are more faithfully modeled as structured collections of multimodal nodes \cite{yan2024graph}. Treating such data as independent, unrelated pairs leaves a substantial portion of the multimodal signal unused, in particular the information carried by the relations among multimodal entities.

\begin{figure}
    \centering
    \includegraphics[width=1\linewidth]{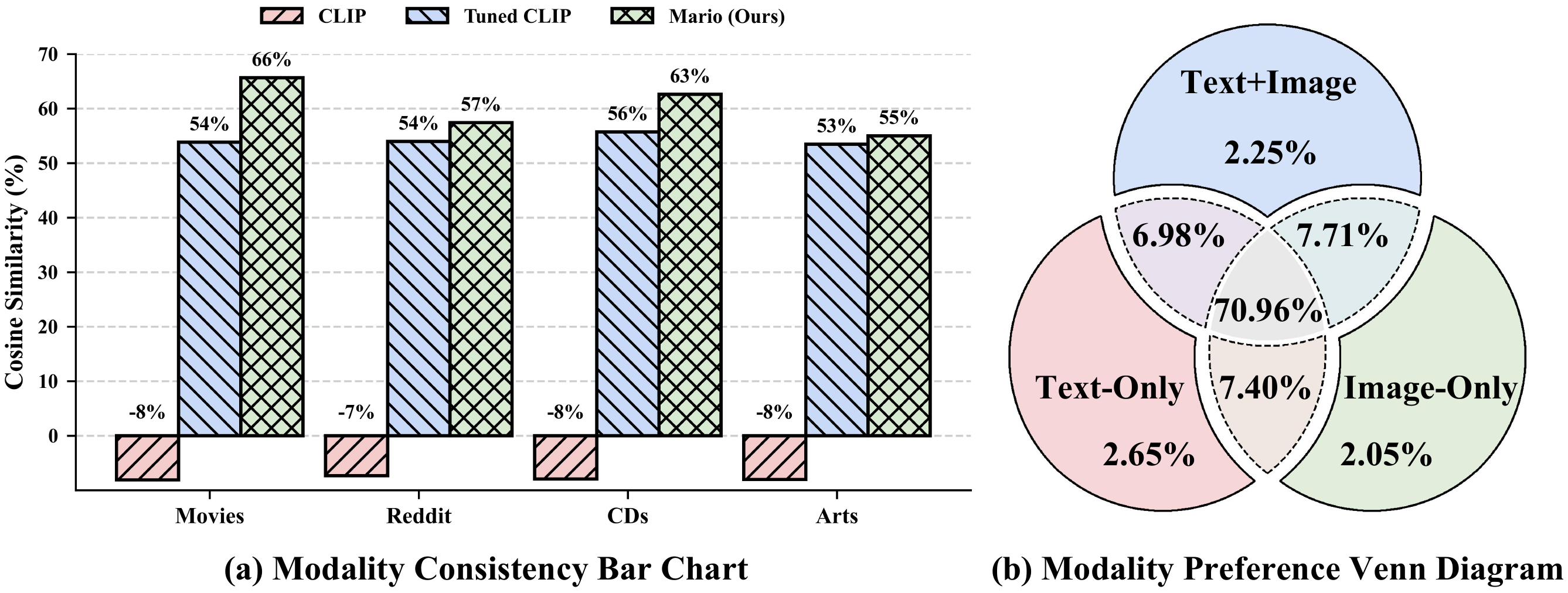}
    \caption{(a) Cosine similarity between text and image embeddings across three models on four datasets.
(b) Venn diagram over three prompt templates with different modality inputs: Text-Only, Image-Only, and Text+Image. Each colored circle corresponds to one template. Among nodes correctly classified by at least one template, each region shows the proportion exclusively covered by the corresponding template combination (where overlapping regions blend the colors). Results are averaged over four datasets.}
\vspace{-0.6cm}
    \label{Fig1}
\end{figure}

This mismatch has prompted a recent line of work to make the relational nature of multimodal data explicit by casting them as multimodal graphs (MMGs) \cite{yan2024graph, zhu2024multimodal,liu2025graph}: each node carries textual and visual attributes while edges supply additional structural priors. A seemingly natural recipe \cite{liu2025graph} then suggests itself: reuse powerful vision–language models (VLMs) (e.g., CLIP \cite{radford2021learning}) to encode the node-level modalities, and hand these embeddings to a graph-capable backbone like GNNs \cite{xu2018powerful,li2018deeper,oono2019graph, kipf2016semi} for propagation and reasoning. This strategy is appealing because it preserves the representational strength of pretrained VLMs and delegates only the structural part to the graph model. Nonetheless, it quietly rests on a strong premise: that the textual and visual views of each node are already semantically well synchronized, and that injecting them into the graph will not magnify any cross-modal mismatch.

However, in realistic MMGs the image attached to a node is not always a clean visual rendering of the text, and the text is not always a faithful caption of the image; in isolation, either side can be short, noisy, or semantically underspecified. This gives rise to our first challenge in multimodal graph reasoning: \textbf{\textit{C1: Weak cross‑modal consistency.}} In such cases, neighboring nodes often provide critical cues that disambiguate modality semantics or reinforce missing information, especially when modalities only partially overlap. As shown in Fig.~\ref{Fig1}'s bar chart, by incorporating graph topology into the alignment process, our model Mario achieves significant cross-modal consistency, yielding a \textbf{68\%} average gain over frozen CLIP and an additional \textbf{6\%} improvement over node-wise fine-tuning. This empirically supports our claim that the design of a structure-aware vision–language model is a necessary prerequisite for LLMs to perform reliable reasoning on multimodal graphs.

When such aligned multimodal features are in place, a second difficulty surfaces—\textbf{\textit{C2: heterogeneous modality preference.}} In conventional
LLM-based Graph Model (GraphLLM) settings \cite{tang2023graphgpt, liu2024can}, where node inputs are unimodal, it is reasonable to employ a shared instruction template for all nodes. However, this assumption fails in MMGs, where the informativeness of each modality can vary significantly across nodes, and the above alignment mechanisms largely focus on common cross-modal patterns, whereas the aligned visual and textual features still retain their individual parts \cite{yuksekgonul2022and}. Some nodes are richly described and thus text-salient, others have noisy text and must rely on distinctive visual cues, and many actually require complementary evidence from both modalities. Moreover, when reasoning over local subgraphs, the “effective” modality for the anchor node can be perturbed by its neighbors’ noisy, incomplete, or redundant modalities. As the Venn diagram in Fig. 1 shows, among nodes correctly classified by at least one template, nearly \textbf{30\%} are covered by only one or two of the three modality-specific templates rather than all three.
This suggests that a one-size-fits-all prompting strategy underutilizes available multimodal supervision, highlighting the need for adaptive and node-specific prompting strategies. This leads us to ask:

\vspace{-0.2cm}
\begin{tcolorbox}[
    colback=gray!10,      
    colframe=black,     
    arc=8pt,              
    boxrule=0.8pt,        
    left=8pt,right=8pt,   
    top=6pt,bottom=6pt    
]
\textit{\textcolor{red}{\textbf{(Q)}} Can we exploit graph structure to enforce reliable cross-modal alignment, and, on top of that, use the aligned representations to drive a dynamic prompting policy that adaptively chooses the most informative modality for each node and its local context?}
\end{tcolorbox}
\vspace{-0.2cm}
To answer this question, we introduce \textbf{Mario}—a dual‑stage \textbf{M}odality‑\textbf{A}daptive \textbf{R}easoning over mult\textbf{i}m\textbf{o}dal graphs framework that tightly couples structure‑aware cross‑modal alignment with instruction‑tuned LLMs. Stage 1 serves as a graph-conditioned vision–language model: it employs a dual-tower encoder augmented with a topology-aware multimodal mixer inspired by GNN‑nested designs \cite{yang2021graphformers} to align text and image features under multi-hop structural guidance, yielding structure-aware, cross-modally coherent node representations.  
Stage 2 then organizes these aligned features into a family of multimodal instruction views and jointly trains a Modality-Adaptive Prompt Router with the LLM, so that for each node Mario can use it to decide which modality configuration to surface to the LLM and route the instance to the most predictive view during inference.
Our key contributions are summarized as follows:

\begin{itemize}
\item We undertake the study of employing LLM reasoning on MMGs, identifying two \textbf{\textit{previously underexplored}} obstacles, cross-modal inconsistency and heterogeneous modality preference, and introduce a novel framework-Mario that simultaneously addresses both challenges.

\item We introduce a graph-conditioned vision–language model, \textbf{\textit{a new VLM paradigm}} that aligns image and text under topology to yield symmetric, structure-aware node representations jointly grounded in both modalities.

\item We break the prevailing reliance of GraphLLMs on a fixed-modality template by introducing modality-adaptive graph instruction tuning, \textbf{\textit{a new tuning scheme}} which routes each node to the most informative modality.

\item We conduct extensive evaluations across diverse MMG benchmarks, demonstrating Mario’s state-of-the-art performance in node classification and link prediction. Notably, Mario consistently outperforms leading baselines, achieving up to \textit{\textbf{1.6×}} gains in zero-shot transfer settings.
\end{itemize}

\begin{figure*}
    \centering
    \includegraphics[width=1\linewidth]{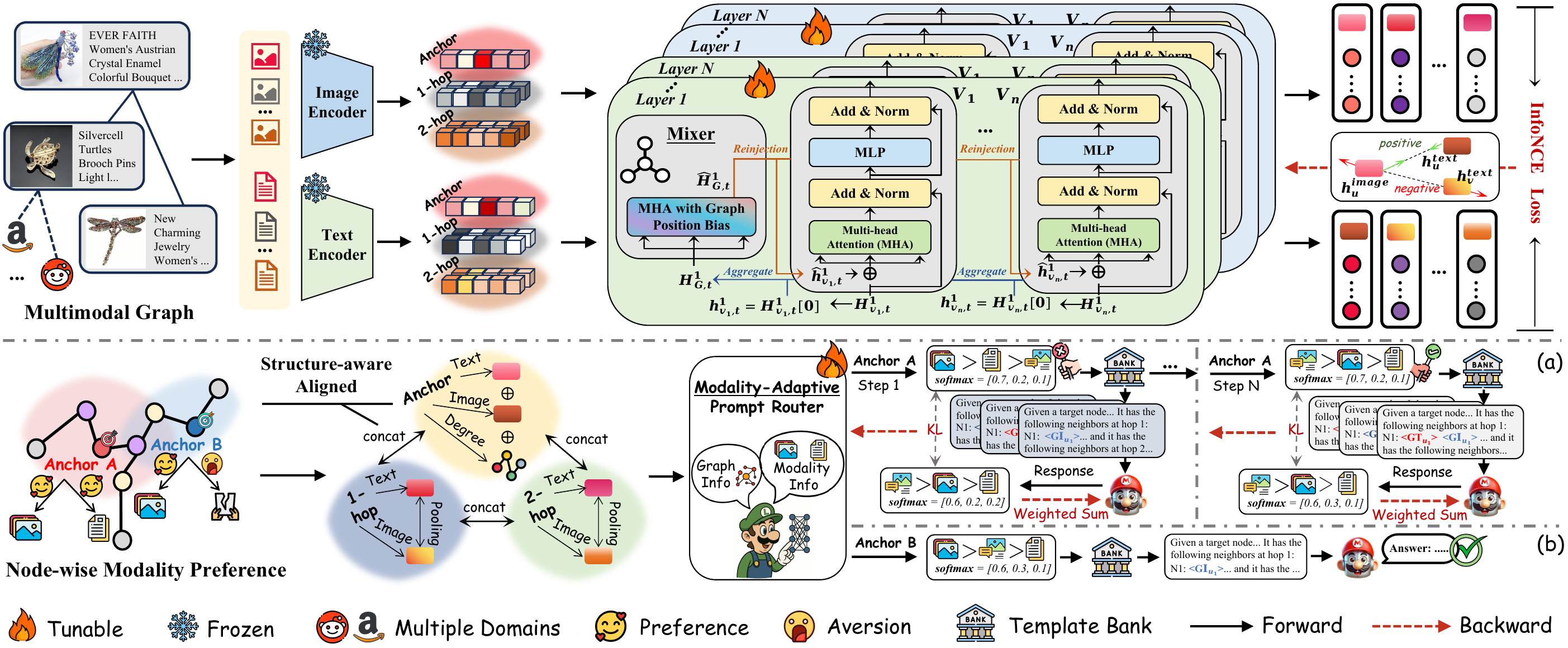}
    \vspace{-0.6cm}
    \caption{Overview of the proposed Mario framework. Given a MMG, Stage 1 uses a graph-conditioned vision–language model to perform structure-aware image–text alignment: images and texts are initially encoded, symmetrically refined by a transformer-embedded mixer that injects graph structure into token embeddings, and then aligned via contrastive learning. Stage 2 builds on these aligned features with modality-adaptive graph instruction tuning, where a lightweight router, trained under LLM supervision (a), infers each node’s modality preference and selects the most suitable modality-specific template for effective multimodal graph reasoning (b).}
    \vspace{-0.6cm}
    \label{fig:pipeline}
\end{figure*}

\section{Related Work}

\textbf{Large Language Models for Graph Reasoning. }
Studies \cite{zhao2023graphtext,guo2023gpt4graph,chen2024exploring, fang2024gaugllm, he2023harnessing} have demonstrated the potential of LLMs in augmenting graph representations and generating contextualized textual information, thereby enhancing semantic expressiveness and improving downstream graph learning performance.
Another line of work utilizes graph or language models to obtain graph tokens, which are then embedded into prompt templates for instruction tuning \cite{tang2023graphgpt, zhang2024graphtranslator, chenllaga, liu2024can} or directly provide LLMs with more text information related to the graph structure through in-context learning (ICL) to directly infer without training \cite{sun2025graphicl, huang2023can}. When applying graph instruction tuning, it enables LLMs to understand these structure-informed tokens and perform appropriate reasoning, leading to better generalization in certain graph-related tasks. However, most of these studies operate only on textual modality (text-attributed graphs) and generalize poorly to multimodal graph reasoning. Moreover, their instruction tuning typically relies on a single template with fixed modality inputs, overlooking that different nodes may favor different modality information.

\noindent\textbf{Multimodal Graph Learning.}
Multimodal graph learning extends conventional graph representation learning by integrating multiple modalities, often text and images, to enhance node/edge representations. While recent multimodal graph datasets and benchmarks \cite{yan2024graph, zhu2024multimodal,liu2025graph}, including some multimodal KG works \cite{huang2025e, liu2025align}, have facilitated research in this area, the development of effective multimodal graph models remains an open challenge.
A common approach \cite{wei2019mmgcn,he2025unigraph2,tao2020mgat} leverages multimodal representation alignment, primarily relying on VLMs \citep{radford2021learning,li2022blip} to generate multimodal node embeddings and do downstream graph tasks. More recently, MLaGA \cite{fan2025mlaga} uses a graph-guided multimodal aligner with instruction tuning to provide LLMs graph-aware multimodal tokens, but it aligns only after fusing text and image into a shared query representation (not per-modality). As a result, it implicitly assumes equal modality utility and fails to address node-level modality inconsistency and heterogeneous modality preference in MMGs. Graph4MM \cite{ning2025graph4mm} introduces hop-diffused attention to inject multi-hop structural information directly into self-attention, but it primarily targets MMGs with missing modalities, overlooking the prevalent fully observed setting where each node has both text and image. Consequently, existing approaches still miss two key realities in MMGs: (i) text–image consistency can be weak, making naïve fusion/alignment unreliable, and (ii) modality preference varies across nodes.

\section{Preliminary}

\noindent\textbf{Multimodal Graph.} A multimodal graph is a structured data format where each node is associated with multiple modalities. In our case, each node has a textual description and an associated image. Formally, the graph is denoted as $\mathcal{G} = (\mathcal{V}, \mathcal{E}, \mathbf{A}, \mathcal{X})$, where $\mathcal{V}$ is the node set, $\mathcal{E}$ the edge set, and $\mathbf{A} \in \mathbb{R}^{N \times N}$ the adjacency matrix with $\mathbf{A}_{ij} = 1$ if $(v_i, v_j) \in \mathcal{E}$ and $0$ otherwise. $\mathcal{X}$ is the set of multimodal node features, where each element $\mathbf{x}_v \in \mathcal{X}$ is a structured pair $(\mathbf{x}_v^{\text{text}}, \mathbf{x}_v^{\text{image}})$, with $\mathbf{x}_v^{\text{text}} \in \mathbb{R}^{d_t}$ and $\mathbf{x}_v^{\text{image}} \in \mathbb{R}^{d_i}$ denoting the textual and visual features of node $v$.

\noindent \textbf{Large Language Models and Instruction Tuning.}
Large Language Models can be adapted to downstream tasks via instruction tuning, which enables parameter-efficient or full updates to model parameters $\theta$. Unlike conventional fine-tuning, instruction tuning enhances LLMs through structured prompts that combine task-specific instructions with optional learnable soft tokens. Given an input sequence of textual tokens $\mathbf{X} = {x_1, x_2, \dots, x_p}$, an instruction function $\mathcal{T}$ produces a structured prompt $\mathcal{T}(\mathbf{X})$. This may be concatenated with a learnable soft token set $\mathbf{S} = {s_1, s_2, \dots, s_q}$ to form the augmented prompt $\mathcal{P}^{} = [\mathcal{T}(\mathbf{X}); \mathbf{S}]$. The model then generates an output sequence $\mathbf{Y}$ conditioned on $\mathcal{P}^{}$ and $\mathbf{X}$. Training jointly optimizes $\mathbf{S}$ and $\theta$ to improve task adaptation while reducing the need for full model retraining.

\vspace{-0.2cm}
\section{Methodology}

In this section, we introduce the detailed architecture of Mario, as illustrated in Figure~\ref{fig:pipeline}. It consists of two novel stages: a graph-conditioned vision–language model (Sec.~\ref{sec:4.1}) that enhances cross-modal consistency in multimodal graphs (\textbf{\textit{C1}} \checkmark), and a modality-adaptive graph instruction tuning scheme (Sec.~\ref{sec:4.2}) tailored to heterogeneous modality preferences (\textbf{\textit{C2}} \checkmark). We conclude with an overview of the model’s training and inference procedures, along with a discussion of runtime complexity (Sec.~\ref{sec:4.3}).

\subsection{Graph-conditioned Vision-Language Model}
\label{sec:4.1}

\textbf{Objective.}\;
Given a multimodal graph $\mathcal G$ where every
node $v$ owns a text sequence~$\mathbf T_v$ and an image
patch sequence~$\mathbf I_v$, Stage~1 learns a latent
space in which \textit{(i)} text and vision cues of the \emph{same} node are
metrically close, \textit{(ii)} \textit{structure-aware alignment} is performed to preserve fine-grained modality information and \textit{(iii)} embeddings respect \emph{neighborhood dependencies} to improve modality consistency. We introduce a graph-conditioned vision--language model
(GVLM) that employs a dual-tower architecture, where each encoder is equipped with a Topology-Aware Multimodal Mixer so that its \texttt{[CLS]} summary attends over all tokens and patches under graph guidance. Bidirectional InfoNCE~\cite{oord2018cpc} is
then applied to these graph-conditioned \texttt{[CLS]} representations, aligning the two streams in a structure-aware manner.

\noindent\textbf{Modality Encoding.}\;
We employ two separate $L$-layer Transformers, one for each modality. For a given node $v$, the hidden representation at layer $l$ is denoted as $\mathbf{H}^l_{v,M} \in \mathbb{R}^{t_M \times d}$, where $M \in {t, i}$ indicates the modality (text or image), $t_t = m$ is the number of text tokens, and $t_i = n$ is the number of image patches. The initial Layer-0 embeddings are derived from pretrained vision-language representations like CLIP, enriched with position embeddings to support subsequent transformer layers.
The \texttt{[CLS]} token embedding,
$\mathbf{h}^l_{v,M}=\mathbf{H}^l_{v,M}[0]$,
serves as the node representation used in the following topology‑aware multimodal mixer.

\noindent\textbf{Topology‑Aware Multimodal Mixer.}\;
For modality $M\!\in\!\{t,i\}$ and layer $l$, we first gather the
\texttt{[CLS]} summaries of all graph nodes to build a
node‑by‑feature matrix:
\begin{equation}
\mathbf H^{l}_{\mathcal G,M}
=\bigl[\mathbf h^{\,l}_{u,M}\bigr]_{u\in\mathcal V}
\in\mathbb R^{|\mathcal V|\times d}.
\end{equation}
Each attention head $h{=}1,\dots,H$ projects the node representations into queries, keys, and values via:
\begin{equation}
\small
\mathbf Q_{M,h},\, \mathbf K_{M,h},\, \mathbf V_{M,h}
= \mathbf H^{l}_{\mathcal G,M} \cdot \left(\mathbf W^Q_{M,h},\mathbf W^K_{M,h},\mathbf W^V_{M,h}\right),
\end{equation}
where each $\mathbf W^{\cdot}_{M,h} \in \mathbb R^{d \times (d/H)}$ is a trainable projection matrix for head $h$.
Scaled dot‑product attention, enriched with a graph‑aware position bias,
is then applied:

\begin{equation}
\widehat{\mathbf H}^{\,l}_{\mathcal G,M}
=\bigl\lVert_{h=1}^{H}
\mathrm{softmax}\!\Bigl(
\tfrac{\mathbf Q_{M,h}\mathbf K_{M,h}^{\!\top}}{\sqrt{d/H}}
+\mathbf B_h\Bigr)\mathbf V_{M,h}.
\end{equation}

The learned position bias $\mathbf B_h$ encodes graph structural roles, distinguishing relationships between the nodes, serving as a form of relative positional information, and is implemented as head-specific learnable scalars indexed by shortest-path-distance buckets.
Concatenating all heads yields
$\widehat{\mathbf H}^{\,l}_{\mathcal G,M}\!\in\!\mathbb
R^{|\mathcal V|\times d}$; where the $v$-th row $\widehat{\mathbf h}^{\,l}_{v,M}$ encodes the structure-aware representation for node $v$

\noindent\textbf{Reinjection for Multimodal Context Integration.}\;
To propagate the structure-aware representation
$\widehat{\mathbf h}^{\,l}_{v,M}$ back into the token stream, we append it
to the front of the modality-specific sequence to replace the previous \texttt{[CLS]} token embedding,
thereby keeping the sequence length and embedding
dimension unchanged.
This augmented sequence is then processed by the next Transformer block:
\begin{equation}
\widetilde{\mathbf H}^{\,l}_{v,M}
= \bigl[\, \widehat{\mathbf h}^{\,l}_{v,M} \;\Vert\;
\mathbf H^{\,l}_{v,M}[1:]\, \bigr].
\end{equation}
A new \texttt{[CLS]} token is produced at each layer, allowing the model
to iteratively refine its node-level representation by blending freshly
aggregated graph context with the original token features.
Repeating this mixer–reinjection operation for $L$ layers yields final, structure-aware
embeddings
\begin{equation}
\mathbf h^{\text{text}}_v=\mathbf H^{L}_{v,t}[0],
\qquad
\mathbf h^{\text{image}}_v=\mathbf H^{L}_{v,i}[0],
\end{equation}
which capture both modality-specific nuances and topology-aware
signals, and will later serve as GVLM prototypes for cross-modal
contrastive alignment.

\noindent\textbf{Cross‑Modal Contrastive Learning.}
To further tighten the gap between modalities, we perform a bidirectional
contrastive objective on the structure‑aware modality embeddings obtained
above to train our GVLM: for each node $v$ in a batch $\mathcal B$, its text–image pair
$\bigl(\mathbf h^{\text{text}}_{v},\mathbf h^{\text{image}}_{v}\bigr)$ is the
sole positive, while all cross‑node combinations serve as negatives sampled all in batch.
Since these embeddings already fold in topology, the neighbor signals—absent in plain text corpora—become the key to narrowing the cross‑modal gap, forcing the model to learn representations that are simultaneously modality‑aligned and structure‑aware.
We minimize the symmetric, temperature‑scaled InfoNCE loss
\begin{equation}
\mathcal L_{\text{S1}}
=-\frac{1}{|\mathcal B|}\sum_{v\in\mathcal B}\!
\bigl[
\log\tfrac{e^{s(v,v)/\tau}}{\sum_{u\in\mathcal B}e^{s(v,u)/\tau}}
+
\log\tfrac{e^{s(v,v)/\tau}}{\sum_{u\in\mathcal B}e^{s(u,v)/\tau}}
\bigr],
\end{equation}
where $s(u,v)$ is the cosine similarity between $\mathbf h^{\text{text}}_u$ and $\mathbf h^{\text{image}}_v$ and $\tau$ controls sharpness.
This step delivers modality‑consistent, topology‑aware representations that
feed directly into Stage 2’s adaptive instruction tuning and empirically boost
downstream cross‑modal coherence.

\subsection{Modality-Adaptive Graph Instruction Tuning}

\label{sec:4.2}

\noindent\textbf{Objective.}\;
Building on the features obtained in Stage 1, Stage 2 endows the LLM with node‑level modality adaptivity.
\textit{(i)} For each node, we construct prompts with multimodal graph signals under three modality views (\emph{text}, \emph{image}, \emph{multimodal}) by blending its features with top 1/2-hop neighbors.
\textit{(ii)} A Modality--Adaptive Prompt Router (MAPR), trained with a probability-weighted LLM loss plus a KL term, reallocates probability toward the view that yields the lowest loss, so the LLM can exploit informative modalities while down-weighting noisy ones.

\noindent\textbf{Multimodal Graph‑Contextual Signals.}\;
For every node $v$, we expose the LLM to both its intrinsic multimodal features and the most relevant structural context.
We introduce two special tokens, $\langle\mathrm{GT}_v\rangle$ and $\langle\mathrm{GI}_v\rangle$, which are placed in the prompt to provide the LLM with text and image information about graph node $v$. Their embeddings are obtained by applying a learnable shared projector $\mathbf P$ to the Stage-1 features $\mathbf h_v^{\text{text}}$ and $\mathbf h_v^{\text{image}}$, mapped into the LLM token embedding space.
To enrich the prompt with neighborhood evidence, we examine
$1$‑hop and $2$‑hop neighbors
$\mathcal N^1(v),\mathcal N^2(v)$ from the training set
and select the Top‑$K$ nodes per hop that are relatively important to $v$ based on the cosine similarity between the concatenated embeddings
$[\mathbf h_u^{\text{text}} \Vert \mathbf h_u^{\text{image}}]$ and
$[\mathbf h_v^{\text{text}} \Vert \mathbf h_v^{\text{image}}]$.
For each chosen neighbor $u$ we create and optionally inject
\(
\langle\text{GT}_u\rangle,\langle\text{GI}_u\rangle
\)
based on the anchor node $v$’s modality preference, and attach its label~$\ell_u$ following the ICL paradigm for LLMs (during inference, this entire procedure is restricted to training nodes only; validation/test nodes are never used as in-context exemplars), thereby forming different modality-specific templates.

\noindent\textbf{Prompt Template Bank.}\;
A node’s modality preference is shaped jointly by its own modalities and its local subgraph. To expose complementary MMG evidence under different preferences, we form three modality-specific special-token groups for node $v$:
\[
\small
\begin{aligned}
\mathcal S_v^{\text{txt}} &= \big\{\, \langle \mathrm{GT}_v\rangle \,;\, \langle \mathrm{GT}_{u_1}\rangle,\dots,\langle \mathrm{GT}_{u_K}\rangle \,\big\},\\[3pt]
\mathcal S_v^{\text{vis}} &= \big\{\, \langle \mathrm{GI}_v\rangle \,;\, \langle \mathrm{GI}_{u_1}\rangle,\dots,\langle \mathrm{GI}_{u_K}\rangle \,\big\},\\[3pt]
\mathcal S_v^{\text{mm}}  &= \big\{\, \langle \mathrm{GT}_v\rangle,\langle \mathrm{GI}_v\rangle \,,\,\dots,\langle \mathrm{GT}_{u_K}\rangle,\langle \mathrm{GI}_{u_K}\rangle \,\big\},
\end{aligned}
\]
where $\langle \mathrm{GT}_\cdot\rangle$ and $\langle \mathrm{GI}_\cdot\rangle$ are the graph-text and graph-image tokens defined earlier.
\noindent We then define the prompt for node $v$ as the concatenation of three parts—$\mathcal I$ (task instruction), $r_v$ (anchor node raw text), and $\mathcal S_v^{(k)}$ (modality-specific special tokens)—i.e., $\mathcal P_v^{(k)} = \mathcal I \oplus r_v \oplus \mathcal S_v^{(k)}$, where $k\in\{\text{txt},\text{vis},\text{mm}\}$.
Here, $\mathcal I$ is a concise instruction (e.g., \textsl{“Predict the node category.”}) that specifies the task; $r_v$ passes the anchor node’s raw textual content to preserve fine-grained semantics; and $\mathcal S_v^{(k)}$ supplies the modality-specific graph signals from $v$ and its top 1/2-hop neighbors,  which seamlessly weaves together node‑specific multimodal cues and the most informative structural signals before being routed to the LLM to unlock the in-context learning potential of LLM in the context of multimodal graphs.

\noindent\textbf{Modality–Adaptive Prompt Router.}\;
However, how to determine each node’s modality preference while being minimally affected by poorer modalities and make the LLM update selectively according to its advantageous modality remains open. To this end, we introduce a lightweight yet expressive Modality--Adaptive Prompt Router (MAPR) placed before the LLM. For each node $v$ we concatenate its Stage 1 multimodal embedding, the mean‑pooled 1‑/2‑hop context, and a logarithmic degree term to serve as input: 
\begin{equation}
\small
\mathbf z_v
=\Bigl[
\mathbf h_v^{\text{text}}\,;\,
\mathbf h_v^{\text{image}}\,;\,
\phi^{(1)}(v)\,;\,
\phi^{(2)}(v)\,;\,
\log d_v
\Bigr] \in \mathbb R^{4d + 1},
\end{equation}
\begin{equation}
\small
\phi^{(h)}(v)
=\frac{1}{2|\mathcal N^h(v)|}
\sum_{u\in\mathcal N^h(v)} \bigl(\mathbf h_u^{\text{text}}+\mathbf h_u^{\text{image}} \bigr),
\end{equation}
For node pairs in the link prediction task, we first pool the two nodes and treat them as a single pseudo-node, then construct the input in the same manner as for regular nodes. Afterwards, we select their common neighbors in the template and embed the corresponding tokens.
During training, we expose all templates in the bank to the LLM so that it learns to rank prompts containing different modality information. The lightweight MLP router fuses diverse information into one feature and obtains modality‑selection logits $\mathbf s_v\in\mathbb R^{3}$ that are normalised to routing probabilities $\mathbf p_v=\mathrm{softmax}(\mathbf s_v)=[p_v^{(\text{txt})},p_v^{(\text{vis})},p_v^{(\text{mm})}]^\top$.  For each template, the LLM produces a negative causal language modeling loss through function:
\vspace{-0.1cm}
\begin{equation}
  \ell^{(k)}_v
  = -\sum_{i=1}^{|\mathbf Y_v|}
     \log p_{\theta}\!\bigl(y_i \mid y_{<i}, \mathcal P_v^{(k)}\bigr),
\end{equation}
which we convert into a performance posterior
\(
\mathbf q_v=\mathrm{softmax}\!\bigl(
    -[\ell^{(\text{txt})}_v,\ell^{(\text{vis})}_v,\ell^{(\text{mm})}_v]
\bigr)
\) and use it to guide MAPR updates. Training minimizes the composite loss:
\vspace{-0.2cm}
\begin{equation}
  \mathcal L_{\text{S2}}
  = \frac{1}{|B|}
    \sum_{v \in B}
    \Bigg[
      \sum_{k} q_v^{(k)}\,\ell^{(k)}_v
      + \lambda\,\mathrm{KL}\!\bigl(\mathbf q_v \,\|\, \mathbf p_v\bigr)
    \Bigg].
\end{equation}

where the first term is a performance-weighted objective that routes gradient to each template in proportion to the posterior $\mathbf q_v$ inferred from its loss, while the KL term regularizes the router by matching its predictive distribution $\mathbf p_v$ to $\mathbf q_v$. This teacher–student coupling shifts probability mass toward the lower-loss templates, encouraging the LLM to rely on informative modalities and to down-weight noisy ones. MAPR stabilizes optimization and improves generalization by tailoring supervisory signal to modality preference, attenuating gradients from mismatched modalities.

\begin{table*}[t]\scriptsize
\centering
\caption{Single-Focus performance comparison on four MMG datasets. Red $\uparrow$ number denotes the absolute accuracy gain over each baseline. Since LLaVA 1.5 does not support multiple image inputs, we concatenate images of node pairs and their neighbors into a single canvas before feeding the model to ensure a fair comparison. We fine-tune Qwen2.5-VL by feeding it the anchor node and its neighbors' texts+images. The best baseline in each modality setting is highlighted in bold, and Mario (ours) is shown in the last row with underline.}
\vspace{-0.2cm}
\label{tab:singlefocus}
\scriptsize{
\resizebox{\linewidth}{!}{
\setlength\tabcolsep{2pt}
\renewcommand\arraystretch{1.2}
\setlength{\arrayrulewidth}{0.3mm}
\begin{tabular}{c||cccc||cccc}
\hline
\rowcolor{gray!30}
\multirow{-1}{*}{\centering \textbf{Methods}} 
& \multicolumn{4}{c||}{\textbf{Node Classification  Accuracy (\%)}} 
& \multicolumn{4}{c}{\textbf{Link Prediction Accuracy (\%)}} \\
\hline
\rowcolor{gray!3}
\textbf{Datasets}
& \textbf{Movies} & \textbf{Reddit} & \textbf{CDs} & \textbf{Arts}
& \textbf{Movies} & \textbf{Reddit} & \textbf{CDs} & \textbf{Arts} \\
\hline

\rowcolor[HTML]{D7F6FF}
\multicolumn{9}{c}{\textbf{\textit{Text-only}}}\\
\hline
\rowcolor{gray!3}
GCN            
& $43.77$ \myred{9.86} & $84.29$ \myred{11.01} & $51.44$ \myred{11.99} & $76.92$ \myred{15.21}
& $70.20$ \myred{23.70} & $74.23$ \myred{17.07} & $71.63$ \myred{21.07} & $66.27$ \myred{23.69} \\

GATv2          
& $48.71$ \myred{4.92} & $85.57$ \myred{9.73} & $54.67$ \myred{8.76} & $80.39$ \myred{11.74}
& $72.63$ \myred{21.27} & $70.63$ \myred{20.67} & $73.27$ \myred{19.43} & $69.47$ \myred{20.49} \\

\rowcolor{gray!3}
SAGE           
& $43.17$ \myred{10.46} & $85.64$ \myred{9.66} & $52.16$ \myred{11.27} & $85.26$ \myred{6.87}
& $65.47$ \myred{28.43} & $71.53$ \myred{19.77} & $68.73$ \myred{23.97} & $65.50$ \myred{24.46} \\

LLaMA3-8B     
& $15.50$ \myred{38.13} & $77.23$ \myred{18.07} & $33.52$ \myred{29.91} & $72.42$ \myred{19.71}
& $63.10$ \myred{30.80} & $79.30$ \myred{12.00} & $74.30$ \myred{18.40} & $66.40$ \myred{23.56} \\

GraphGPT  
& $23.85$ \myred{29.78} & $22.99$ \myred{72.31} & $23.85$ \myred{39.58} & $57.35$ \myred{34.78}
& $53.50$ \myred{40.40} & $42.90$ \myred{48.40} & $68.95$ \myred{23.75} & $65.46$ \myred{24.50} \\

\rowcolor{gray!3}
LLaGA          
& $\mathbf{49.57}$ \myred{4.06} & $\mathbf{92.14}$ \myred{3.16} & $\mathbf{54.74}$ \myred{8.69} & $\mathbf{89.32}$ \myred{2.81}
& $\mathbf{75.90}$ \myred{18.00} & $\mathbf{88.45}$ \myred{2.85} & $\mathbf{84.90}$ \myred{7.80} & $\mathbf{82.30}$ \myred{7.66} \\

GraphPrompter  
& $46.35$ \myred{7.28} & $91.16$ \myred{4.14} & $46.27$ \myred{17.16} & $84.41$ \myred{7.72}
& $67.90$ \myred{26.00} & $87.73$ \myred{3.57} & $78.70$ \myred{14.00} & $71.80$ \myred{18.16} \\
\hline

\rowcolor[HTML]{D7F6FF}
\multicolumn{9}{c}{\textbf{\textit{Image-only}}}\\
\hline
\rowcolor{gray!3}
GCN  
& $45.24$ \myred{8.39} & $88.63$ \myred{6.67} & $51.61$ \myred{11.82} & $76.24$ \myred{15.89}
& $69.53$ \myred{24.37} & $\mathbf{75.07}$ \myred{16.23} & $71.43$ \myred{21.27} & $67.47$ \myred{22.49} \\

SAGE   
& $44.37$ \myred{9.26} & $89.83$ \myred{5.47} & $54.94$ \myred{8.49} & $\mathbf{80.29}$ \myred{11.84}
& $73.50$ \myred{20.40} & $74.00$ \myred{17.30} & $67.67$ \myred{25.03} & $62.93$ \myred{27.03} \\

\rowcolor{gray!3}
GATv2   
& $\mathbf{50.02}$ \myred{3.61} & $\mathbf{89.87}$ \myred{5.43} & $\mathbf{55.82}$ \myred{7.61} & $78.46$ \myred{13.67}
& $\mathbf{74.67}$ \myred{19.23} & $74.50$ \myred{16.80} & $\mathbf{73.40}$ \myred{19.30} & $\mathbf{69.37}$ \myred{20.59} \\

LLaVA1.5-13B   
& $17.78$ \myred{35.85} & $65.78$ \myred{29.52} & $30.13$ \myred{33.30} & $52.01$ \myred{40.12}
& $45.50$ \myred{48.40} & $49.80$ \myred{41.50} & $50.13$ \myred{42.57} & $51.10$ \myred{38.86}\\
\hline

\rowcolor[HTML]{D7F6FF}
\multicolumn{9}{c}{\textbf{\textit{Text+Image}}}\\
\hline
\rowcolor{gray!3}
GCN            
& $46.96$ \myred{6.67} & $88.09$ \myred{7.21} & $52.67$ \myred{10.76} & $76.76$ \myred{15.37}
& $69.93$ \myred{23.97} & $74.37$ \myred{16.93} & $71.07$ \myred{21.63} & $67.53$ \myred{22.43} \\

GATv2          
& $49.29$ \myred{4.34} & $89.80$ \myred{5.50} & $56.44$ \myred{6.99} & $81.19$ \myred{10.94}
& $72.73$ \myred{21.17} & $72.67$ \myred{18.63} & $73.20$ \myred{19.50} & $70.03$ \myred{19.93} \\

\rowcolor{gray!3}
SAGE           
& $44.07$ \myred{9.56} & $90.21$ \myred{5.09} & $54.74$ \myred{8.69} & $85.35$ \myred{6.78}
& $70.27$ \myred{23.63} & $72.00$ \myred{19.30} & $68.37$ \myred{24.33} & $66.27$ \myred{23.69} \\

LLaVA1.5-13B   
& $18.89$ \myred{34.74} & $72.33$ \myred{22.97} & $40.10$ \myred{23.33} & $57.63$ \myred{34.50}
& $65.80$ \myred{28.10} & $74.63$ \myred{16.67} & $56.90$ \myred{35.80} & $67.90$ \myred{23.86} \\

\rowcolor{gray!3}
Qwen2.5-VL
& $49.86$ \myred{3.77} & $70.11$ \myred{25.19} & $53.52$ \myred{9.91} & $88.99$ \myred{3.14}
& $88.10$ \myred{5.80} & $88.20$ \myred{3.10} & $80.90$ \myred{11.80} & $80.40$ \myred{9.56} \\

UniGraph2
& $45.91$ \myred{7.72} & $92.65$ \myred{2.65} & $52.13$ \myred{11.30} & $78.81$ \myred{13.32}
& $64.60$ \myred{29.30} & $80.40$ \myred{10.90} & $81.00$ \myred{11.70} & $67.80$ \myred{22.16} \\

\rowcolor{gray!3}
GraphGPT-A  
& $18.81$ \myred{34.82} & $21.98$ \myred{73.32} & $29.56$ \myred{33.87} & $58.73$ \myred{33.40}
& $49.55$ \myred{44.35} & $51.55$ \myred{39.75} & $72.37$ \myred{20.33} & $63.46$ \myred{26.50} \\

LLaGA-A
& $50.61$ \myred{3.02} & $92.94$ \myred{2.36} & $56.29$ \myred{7.14} & $88.83$ \myred{3.30}
& $77.90$ \myred{16.00} & $88.90$ \myred{2.40} & $\mathbf{84.15}$ \myred{8.55} & $82.05$ \myred{7.91} \\

\rowcolor{gray!3}
GraphPrompter-A  
& $45.54$ \myred{8.09} & $92.85$ \myred{2.45} & $52.06$ \myred{11.37} & $83.86$ \myred{8.27}
& $70.90$ \myred{23.00} & $87.82$ \myred{3.48} & $79.83$ \myred{12.87} & $80.11$ \myred{9.85} \\

Graph4MM 
& $\mathbf{51.07}$ \myred{2.56} & $\mathbf{92.89}$ \myred{2.41} & $55.53$ \myred{7.90} & $89.32$ \myred{2.81}
& $\mathbf{90.24}$ \myred{3.66} & $\mathbf{90.80}$ \myred{0.50} & $83.21$ \myred{9.49} & $81.60$ \myred{8.36} \\

\rowcolor{gray!3}
MLaGA  
& $49.42$ \myred{4.21} & $89.79$ \myred{5.51} & $\mathbf{56.45}$ \myred{6.98} & $\mathbf{89.82}$ \myred{2.31}
& $89.96$ \myred{3.94} & $90.35$ \myred{0.95} & $72.87$ \myred{19.83} & $\mathbf{82.97}$ \myred{6.99} \\
\hline

\rowcolor[HTML]{FFF0C1}
Mario-8B (ours)
& $\underline{\mathbf{53.63}}$ & $\underline{\mathbf{95.30}}$ & $\underline{\mathbf{63.43}}$ & $\underline{\mathbf{92.13}}$
& $\underline{\mathbf{93.90}}$ & $\underline{\mathbf{91.30}}$ & $\underline{\mathbf{92.70}}$ & $\underline{\mathbf{89.96}}$ \\
\hline
\end{tabular}}}
\vspace{-0.3cm}
\label{tab:single_focus}
\end{table*}
\subsection{Discussion}
\label{sec:4.3}
\textbf{Training \& Inference Strategy.}
Mario adopts a sequential routine. \textit{(i) Stage1 pre‑training.} We first optimize the GVLM with the cross‑modal InfoNCE loss~$\mathcal L_{\text{S1}}$ until convergence, obtaining $\Theta_{\text{S1}}^{\star}$. \textit{(ii) Stage 2 instruction tuning.}  Keeping $\Theta_{\text{S1}}^{\star}$ fixed, we then fine‑tune the LLM using LoRA \cite{hu2021lora} together with MAPR using the composite loss~$\mathcal L_{\text{S2}}$. The datasets used in the two training stages are kept identical. At inference time, the MAPR switches from the soft routing used in training to a hard policy, selecting the template
$\displaystyle k^{\star}= \arg\max_{} p^{(k)}_{v}$,
and feeds only the corresponding prompt into the LLM, thus incurring no extra compute compared with a single‑template pipeline.

\noindent\textbf{Complexity Analysis.}
\textit{(i) Stage 1.}
Each multimodal mixer layer attends over all nodes’ \texttt{[CLS]} tokens with graph bias, yielding a per‑layer cost
$\mathcal O(|\mathcal V_s|^{2}d)$, where $\mathcal V_s$ denotes the sampled node set and typically $|\mathcal V_s|\ll|\mathcal V|$.
However, only 1–2 layers are sufficient in practice to reach alignment convergence, so the overall training time remains moderate compared with a deeper vanilla Transformer stack.
\textit{(ii) Stage 2.}
For every training sample we execute \emph{three} forward–backward passes—one per template—yielding a cost of
$\mathcal O\!\bigl(3\,f_{\text{LLM}}\bigr)$.
Empirically, the router allows the model to converge in roughly half the epochs needed by a single template baseline with lower losses, offset by the extra per-step computation (see the empirical training curve in Figure~\ref{fig:3a}).

\section{Experiments}
We conducted extensive experiments, primarily aimed at addressing the following research questions (\textbf{\textit{RQs}}): \textit{\textbf{RQ1:}} How does Mario perform on standard multimodal graph reasoning tasks compared to leading baselines that take different modalities as input?  \textit{\textbf{RQ2:}} How well does Mario generalize in zero-shot settings when evaluated on entirely unseen MMGs?  \textit{\textbf{RQ3:}} To what extent does the graph-conditioned vision–language model enhance representation learning and contribute to the multimodal instruction tuning process? \textit{\textbf{RQ4:}} How does Modality–Adaptive Graph Instruction Tuning improve predictive performance over single-template baselines and how efficient is it?

\begin{table*}[t]\scriptsize
\centering
\caption{ Mix-Training performance comparison on four multimodal graph datasets. The table design is consistent with Table ~\ref{tab:singlefocus}.}
\vspace{-0.2cm}
\label{tab:mixtraining}
\scriptsize{
\resizebox{\linewidth}{!}{
\setlength\tabcolsep{2pt}
\renewcommand\arraystretch{1.2}
\setlength{\arrayrulewidth}{0.3mm}
\begin{tabular}{c||cccc||cccc}
\hline
\rowcolor{gray!30}
\multirow{-1}{*}{\centering \textbf{Methods}} 
& \multicolumn{4}{c||}{\textbf{Node Classification Accuracy (\%)}} 
& \multicolumn{4}{c}{\textbf{Link Prediction Accuracy (\%)}} \\
\hline
\rowcolor{gray!3}
\textbf{Datasets}
& \textbf{Movies} & \textbf{Reddit} & \textbf{CDs} & \textbf{Arts}
& \textbf{Movies} & \textbf{Reddit} & \textbf{CDs} & \textbf{Arts} \\
\hline

\rowcolor[HTML]{D7F6FF}
\multicolumn{9}{c}{\textbf{\textit{Text-only}}}\\
\hline
\rowcolor{gray!3}
GCN            
& $47.15$ \myred{3.83} & $86.17$ \myred{7.03} & $50.76$ \myred{9.17} & $79.03$ \myred{12.17}
& $70.30$ \myred{22.15} & $74.50$ \myred{16.23} & $72.93$ \myred{20.57} & $62.33$ \myred{30.27} \\

SAGE           
& $46.85$ \myred{4.13} & $89.96$ \myred{3.24} & $\mathbf{53.24}$ \myred{6.69} & $\mathbf{87.46}$ \myred{3.74}
& $63.30$ \myred{29.15} & $72.03$ \myred{18.70} & $62.03$ \myred{31.47} & $63.70$ \myred{28.90} \\

\rowcolor{gray!3}
LLaGA          
& $\mathbf{47.80}$ \myred{3.18} & $\mathbf{91.14}$ \myred{2.06} & $51.33$ \myred{8.60} & $74.02$ \myred{17.18}
& $\mathbf{87.28}$ \myred{5.17} & $\mathbf{88.95}$ \myred{1.78} & $\mathbf{90.32}$ \myred{3.18} & $87.06$ \myred{5.54} \\

GraphPrompter  
& $45.21$ \myred{5.77} & $90.36$ \myred{2.84} & $44.70$ \myred{15.23} & $83.91$ \myred{7.29}
& $70.27$ \myred{22.18} & $86.70$ \myred{4.03} & $83.20$ \myred{10.30} & $\mathbf{88.30}$ \myred{4.30} \\

LLaMA3-8B      
& $33.27$ \myred{17.71} & $59.63$ \myred{33.57} & $35.76$ \myred{24.17} & $65.21$ \myred{25.99}
& $63.10$ \myred{29.35} & $69.51$ \myred{21.22} & $74.10$ \myred{19.40} & $67.40$ \myred{25.20} \\
\hline

\rowcolor[HTML]{D7F6FF}
\multicolumn{9}{c}{\textbf{\textit{Image-only}}}\\
\hline
\rowcolor{gray!3}
GCN            
& $\mathbf{39.20}$ \myred{11.78} & $89.30$ \myred{3.90} & $51.20$ \myred{8.73} & $75.03$ \myred{16.17}
& $\mathbf{66.17}$ \myred{26.28} & $\mathbf{70.83}$ \myred{19.90} & $\mathbf{67.07}$ \myred{26.43} & $\mathbf{62.00}$ \myred{30.60} \\

SAGE            
& $38.74$ \myred{12.24} & $\mathbf{90.25}$ \myred{2.95} & $\mathbf{53.60}$ \myred{6.33} & $\mathbf{79.40}$ \myred{11.80}
& $51.70$ \myred{40.75} & $68.20$ \myred{22.53} & $62.93$ \myred{30.57} & $59.37$ \myred{33.23} \\
\hline

\rowcolor[HTML]{D7F6FF}
\multicolumn{9}{c}{\textbf{\textit{Text+Image}}}\\
\hline
\rowcolor{gray!3}
GCN            
& $47.76$ \myred{3.22} & $89.87$ \myred{3.33} & $52.28$ \myred{7.65} & $78.59$ \myred{12.61}
& $71.70$ \myred{20.75} & $74.30$ \myred{16.43} & $68.60$ \myred{24.90} & $66.87$ \myred{25.73} \\

SAGE           
& $47.75$ \myred{3.23} & $90.39$ \myred{2.81} & $56.51$ \myred{3.42} & $\mathbf{88.09}$ \myred{3.11}
& $64.30$ \myred{28.15} & $71.23$ \myred{19.50} & $68.43$ \myred{25.07} & $61.33$ \myred{31.27} \\

\rowcolor{gray!3}
MLaGA            
& $\mathbf{50.08}$ \myred{0.90} & $\mathbf{91.45}$ \myred{1.75} & $\mathbf{57.46}$ \myred{2.47} & $86.32$ \myred{4.88}
& $\mathbf{90.90}$ \myred{1.55} & $\mathbf{90.56}$ \myred{0.17} & $\mathbf{86.53}$ \myred{6.97} & $\mathbf{83.57}$ \myred{9.03} \\
\hline

\rowcolor[HTML]{FFF0C1}
Mario-8B (ours)
& $\underline{\mathbf{50.98}}$ & $\underline{\mathbf{93.20}}$ & $\underline{\mathbf{59.93}}$ & $\underline{\mathbf{91.20}}$
& $\underline{\mathbf{92.45}}$ & $\underline{\mathbf{90.73}}$ & $\underline{\mathbf{93.50}}$ & $\underline{\mathbf{92.60}}$ \\
\hline
\end{tabular}}}
\vspace{-0.4cm}
\label{tab:mix_training}
\end{table*}

\subsection{Experiment Configurations}
\textbf{Datasets.}
Our experiments span a diverse range of domains for MMGs, including
\emph{E-commerce}: Amazon-Arts\&Crafts \cite{hou2024bridging}, Amazon-CDs\&Vinyl \cite{hou2024bridging}, Amazon-Toys \cite{yan2024graph}, Amazon-Movies \cite{yan2024graph},
\emph{Social networks}: Reddit-S \cite{yan2024graph}, and
\emph{Literature}: Goodreads (Books) \cite{zhu2024multimodal}.
In these graphs, nodes represent items or posts, and edges indicate co-purchase or co-comment relationships.
Among them, four datasets are used in the Single-Focus (Table~\ref{tab:singlefocus}) and Mix-Training (Table~\ref{tab:mixtraining})  settings, while the remaining two are reserved for transfer experiments (Table~\ref{tab:cross_dataset}), covering two primary tasks: node classification (NC) and link prediction (LP). Appendix shows the details of datasets and  splits.

\noindent\textbf{Baselines.}
To ensure fair and comprehensive comparison, we categorize baselines by input modality:
(1)~\textbf{Text-only:}
Text-based GNNs (GCN \cite{kipf2016semi}, GraphSAGE~\cite{li2021training}, GATv2~\cite{brody2021attentive}), LLM (LLaMA3-8B with LoRA), and text-centric GraphLLMs (LLaGA \cite{chen2024llaga}, GraphPrompter~\cite{liu2024can}, GraphGPT~\cite{tang2023graphgpt}).
(2)~\textbf{Image-only:}
GNNs over image embeddings and LVLMs such as LLaVA v1.5-13B \cite{liu2024visual} limited to visual inputs.
(3)~\textbf{Text+Image:}
GNNs on fused text–image features (Graph-MLLM Style \cite{liu2025graph}), MLaGA \cite{fan2025mlaga}, Graph4MM \cite{ning2025graph4mm} and UniGraph2 \cite{he2025unigraph2}; LVLMs (Tuned Qwen2.5-VL \cite{bai2025qwen25vl}) with dual-modality inputs; and augmented text-based GraphLLMs (the “-A” variant) that simulate multimodal reasoning by appending image captions to textual inputs to help them do multimodal reasoning. 

\noindent\textbf{Implementation details.} 
Our default backbone LLM is LLaMA3.1-8B.
As shown in our backbone ablation results (Appendix), Mario is largely robust to the choice of LLM backbone. 
We by default adopt CLIP encoders to extract initial node embeddings from textual and visual modalities, as it yields satisfactory performance in our preliminary evaluations without further fine-tuning. 
\textbf{\textit{In addition to all experiments presented below, we also provide in the Appendix more details on:}} hyperparameter settings, LLM backbone ablations, variance analysis, t-SNE analysis of our GVLM alignment, comparisons between frozen and LoRA-tuned Mario, template design, GPU resources, as well as additional main experiment results and sensitivity analyses. All results are averaged over three runs.

\subsection{Overall Performance Comparison (RQ1)}

We benchmark Mario on NC and LP, training and testing on each dataset separately. We refer to this setting as the Single-Focus regime. Table~\ref{tab:single_focus} reports results against baselines categorized by their underlying modality.

\textit{\textbf{\underline{Observation 1:} Mario delivers the highest accuracy across all datasets and both tasks under the single-focus setting.}}  
For instance, it lifts NC performance on CDs from \textbf{56.45\%} (best baseline) to \textbf{63.43\%}, and LP accuracy improves by an average of \textbf{4.73\%} across four datasets.  
These gains stem from the synergy between \emph{Stage 1}’s structure‑aware image–text alignment and \emph{Stage 2}’s modality‑adaptive graph instruction tuning, which together encode product semantics and relations more faithfully.

\textit{\textbf{\underline{Observation 2:} Enabling LLMs to directly interpret aligned multimodal features with structural information is more effective than augmenting input via image-to-text conversion.}} 
Augmented GraphLLMs (e.g.\ LLaGA, GraphPrompter, GraphGPT) relatively trail Mario by an average  \textbf{5.48\%}, \textbf{11.00\%} and \textbf{135.9\%}, respectively, on NC.  
Retaining modality‑specific vectors—augmented by graph context—preserves fine‑grained semantics and avoids the noise and redundancy introduced by text‑only surrogates.

\subsection{Generalization \& Transferability (RQ2)}

We first train Mario on an equal four‑way mixture of datasets and test on each domain individually (Table~\ref{tab:mix_training}) and refer to this as data generalization.  
Next, we evaluate \emph{zero‑shot transfer}: the model is trained on one (or several) source graphs and assessed on an unseen graph (Table~\ref{tab:cross_dataset}).

\textit{\textbf{\underline{Observation 3:} Although the model's performance declines to some extent under the Mix-Training setting, Mario manages to maintain a relatively small performance drop and even works better while still preserving a significant lead over the baselines.}} In Table ~\ref{tab:mix_training}, Mario achieves an average relative improvement of \textbf{2.88\%} in NC and \textbf{2.57\%} in LP over the best baseline, further proving Mario's strong generalization across multiple datasets in joint training, maintaining superior performance while adapting to diverse graph structures and modalities.

\begin{table}[ht]
\vspace{-5pt}
\centering
\caption{Zero-Shot Results (Accuracy).}
\vspace{-0.2cm}
\small
\resizebox{\columnwidth}{!}{%
    \setlength\tabcolsep{14pt}
    \renewcommand\arraystretch{1.2}
    \begin{tabular}{c||cc||cc||cc}
        \hline \thickhline
        \rowcolor{CadetBlue!20}
        & \multicolumn{2}{c||}{\textbf{Toys $\rightarrow$ Movies}} 
        & \multicolumn{2}{c||}{\textbf{Toys+Movies $\rightarrow$ CDs}} 
        & \multicolumn{2}{c}{\textbf{Toys $\rightarrow$ Books}} \\
        \cline{2-7}
        \rowcolor{CadetBlue!20}
        \multirow{-2}{*}{\begin{tabular}{c}\textbf{Model}\\\textbf{Tasks}\end{tabular}}
        & \textbf{NC} & \textbf{LP} & \textbf{NC} & \textbf{LP} & \textbf{NC} & \textbf{LP} \\
        \hline\hline
        GCN           & $5.29$  & $62.23$ & $5.64$  & $55.70$ & $11.47$ & $45.43$ \\
        GATv2         & $6.58$  & $62.03$ & $6.58$  & $63.00$ & $14.50$ & $49.00$ \\
        SAGE          & $3.61$  & $61.80$ & $5.87$  & $66.13$ & $7.73$  & $48.40$ \\
        GraphPrompter & $11.15$ & $72.70$ & $36.66$ & $51.00$ & $23.00$ & $62.50$ \\
        LLaGA         & $8.63$  & $62.00$ & $9.69$  & $72.85$ & $8.85$  & $52.50$ \\
        MLaGA         & $24.95$ & $79.87$ & $16.20$ & $52.82$ & $44.85$      & $57.85$ \\
        \rowcolor[HTML]{D7F6FF}
        Mario-8B (Ours) 
                      & $\mathbf{41.00}$ & $\mathbf{86.60}$ 
                      & $\mathbf{54.32}$ & $\mathbf{82.50}$ 
                      & $\mathbf{47.30}$ & $\mathbf{78.30}$ \\
        \hline
    \end{tabular}%
}
\label{tab:cross_dataset}
\vspace{-0.3cm}
\end{table}

\textit{\textbf{\underline{Observation 4:} Mario achieves robust zero-shot reasoning in multimodal graph inference, outperforming baselines by a notable margin.}} Mario achieves \textbf{1.64$\times$} higher NC accuracy than the best baseline on \textit{Toys → Movies}, 
\textbf{1.48$\times$} on \textit{Toys+Movies → CDs}, and \textbf{1.25$\times$} higher LP accuracy on \textit{Toys → Movies}. This can be attributed to its GVLM, which preserves graph-invariant semantics across modalities, and the modality-adaptive router, which induces transferable inductive bias by dynamically selecting the most informative prompt per node—even in unseen graph topologies.
These results highlight Mario’s strong zero-shot reasoning ability across diverse unseen domains.

\subsection{Ablation Study (RQ3)}
To address RQ3, we replace Stage 1's GVLM with other
architectures that capture graph structural information, such as GNNs. We evaluate these models
on NC across three datasets to validate the superiority of our GVLM design.
\begin{figure}[t]
    \centering
    \begin{subfigure}[b]{0.48\linewidth}
        \includegraphics[width=\linewidth]{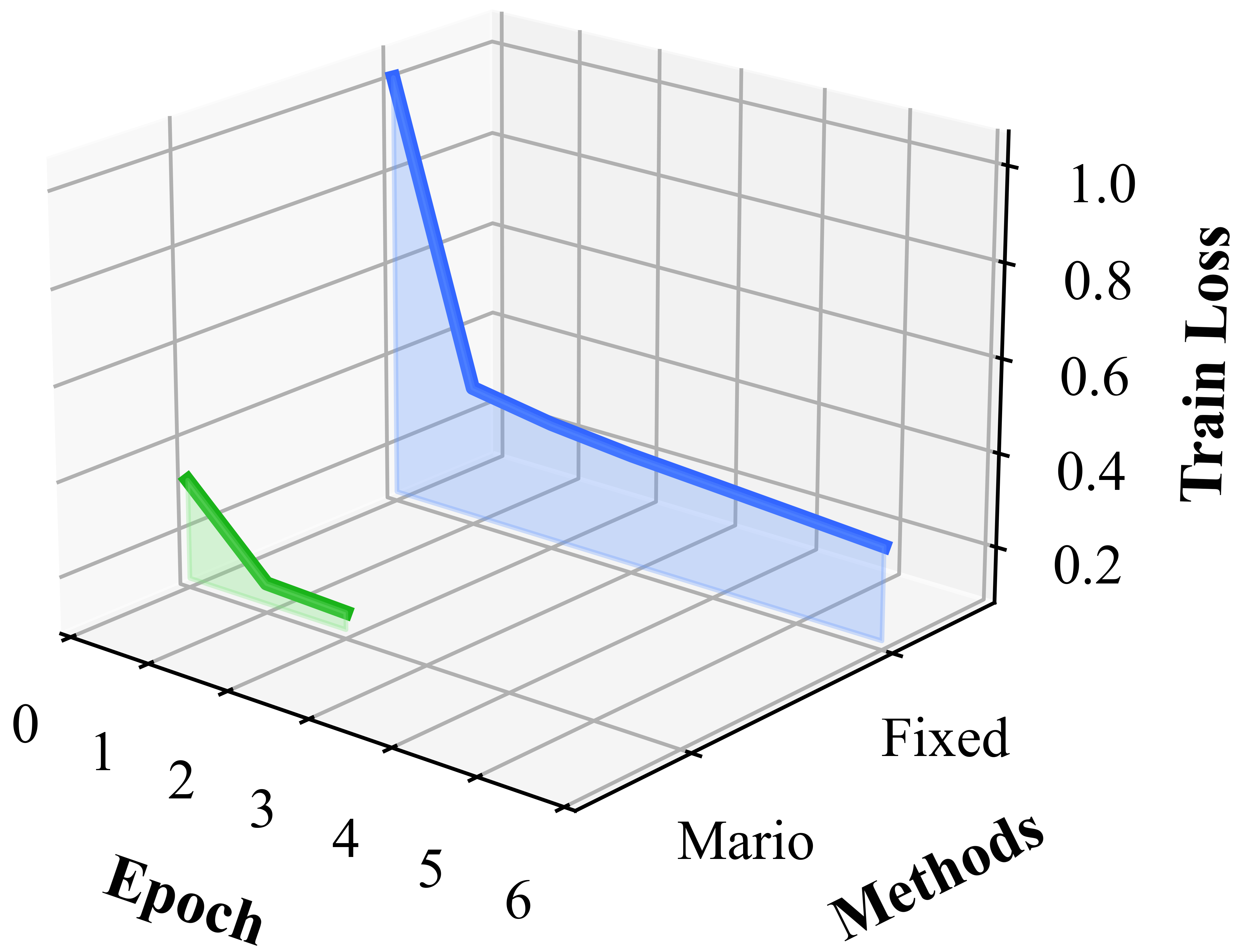}
        \caption{Movies}
        \label{fig:2a}
    \end{subfigure}
    \hfill
    \begin{subfigure}[b]{0.48\linewidth}
        \includegraphics[width=\linewidth]{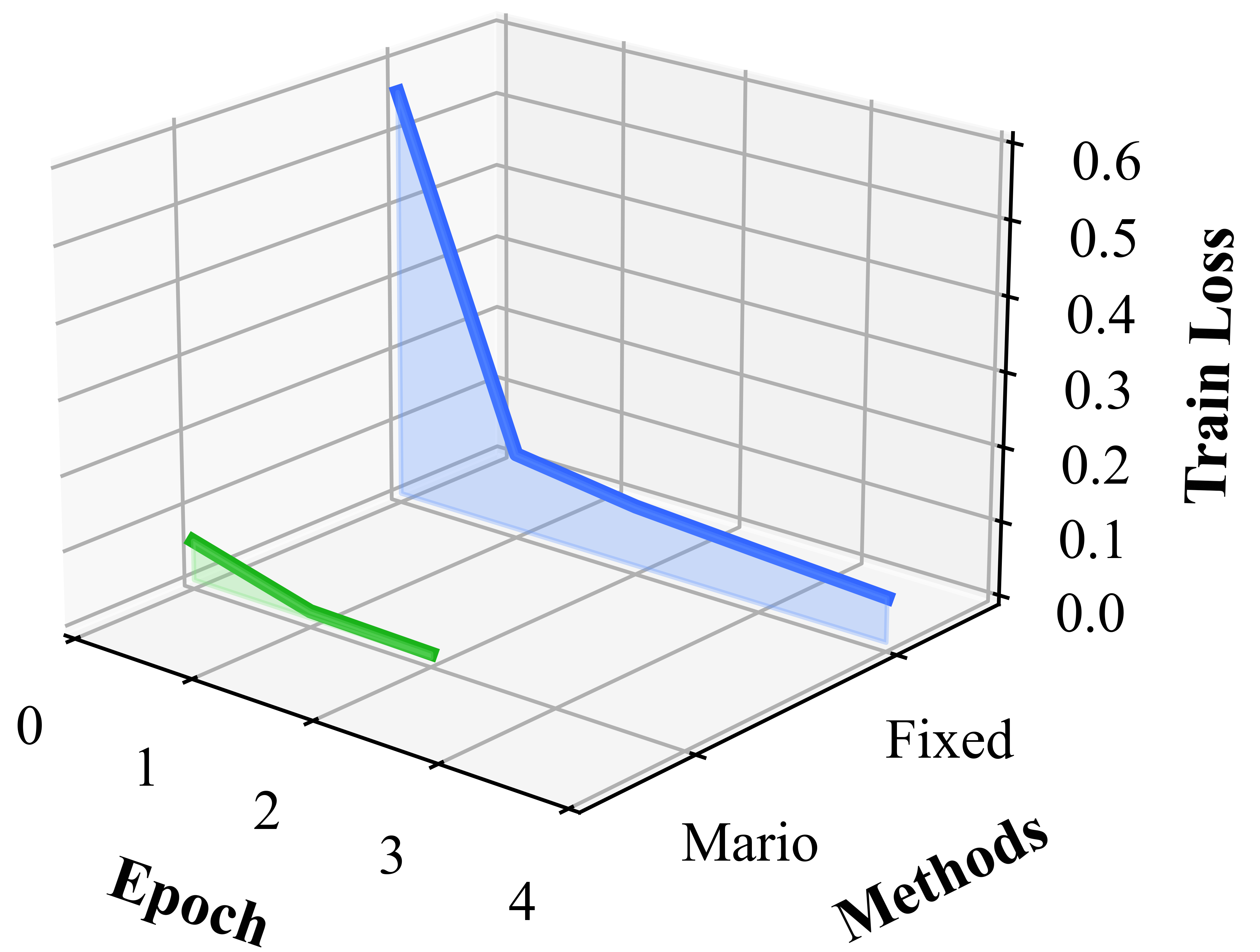}
        \caption{Reddit}
        \label{fig:22}
    \end{subfigure}
    \vspace{-0.2cm}
    \caption{Training curves of Mario vs. the text-only template (Fixed) on two datasets, with early-stopping epochs in the end.}
    \vspace{-0.6cm}
    \label{fig:3a}
\end{figure}

\vspace{-0.4cm}

\textit{\textbf{\underline{Observation 5:} Fine-grained alignment of structured image-text features leads to stronger LLM reasoning than global or structure-agnostic alignment.}}
GNNs and MLP often overlook token-level interactions, losing fine visual details. As shown in Table~\ref{tab:stage1_results}, GVLM surpasses GNNs and MLP across all datasets, especially achieving up to \textbf{+5.15\%} relative average gain on Movies.
While vanilla Transformer-based models are more complex, GVLM adopts a shallow design and converges quickly. Its runtime is only 1.5× that of GNNs/MLPs—an acceptable trade-off for significantly better performance.
\begin{table}[htbp]
    \centering
    \vspace{-0.3cm}
    \caption{Ablation Study of Stage~1's Model.}
    \vspace{-0.2cm}
    \scriptsize
    \setlength{\tabcolsep}{2.5pt}
    \renewcommand{\arraystretch}{1.0}
    \vspace{-0.2cm}
    \begin{tabular}{c||cc|cc|cc}
        \hline\hline
        \rowcolor{CadetBlue!20}
        \textbf{Model} 
        & \multicolumn{2}{c|}{\textbf{Arts}} 
        & \multicolumn{2}{c|}{\textbf{Reddit}} 
        & \multicolumn{2}{c}{\textbf{Movies}} \\
        \rowcolor{CadetBlue!20}
        & Acc(\%) & s/epoch & Acc(\%) & s/epoch & Acc(\%) & s/epoch \\
        \hline\hline
        GCN  
        & $90.32$ \myred{2.00} & $114$  
        & $93.30$ \myred{2.14} & $85$ 
        & $50.90$ \myred{5.36} & $78$ \\
        SAGE 
        & $90.75$ \myred{1.52} & $117$  
        & $92.10$ \myred{3.47} & $89$ 
        & $51.10$ \myred{4.95} & $81$ \\
        GATv2 
        & $90.03$ \myred{2.33} & $129$  
        & $92.90$ \myred{2.58} & $93$ 
        & $51.30$ \myred{4.54} & $85$ \\
        MLP  
        & $89.95$ \myred{2.42} & $109$  
        & $92.70$ \myred{2.80} & $79$ 
        & $50.70$ \myred{5.78} & $75$ \\
        \hline
        \rowcolor[HTML]{D7F6FF}
        \textbf{GVLM} 
        & \textbf{$92.13$} & $174$ 
        & \textbf{$95.30$} & $135$ 
        & \textbf{$53.63$} & $122$ \\
        \hline
    \end{tabular}
    \vspace{-0.2cm}
    \label{tab:stage1_results}
\end{table}
\vspace{-0.2cm}
\subsection{Efficiency Study and Visualization (RQ4)}
In this section, we delve into the effectiveness and efficiency of Modality-Adaptive Graph Instruction Tuning.

\begin{figure}[t]
    \centering
    \begin{subfigure}[b]{0.48\linewidth}
        \includegraphics[width=\linewidth]{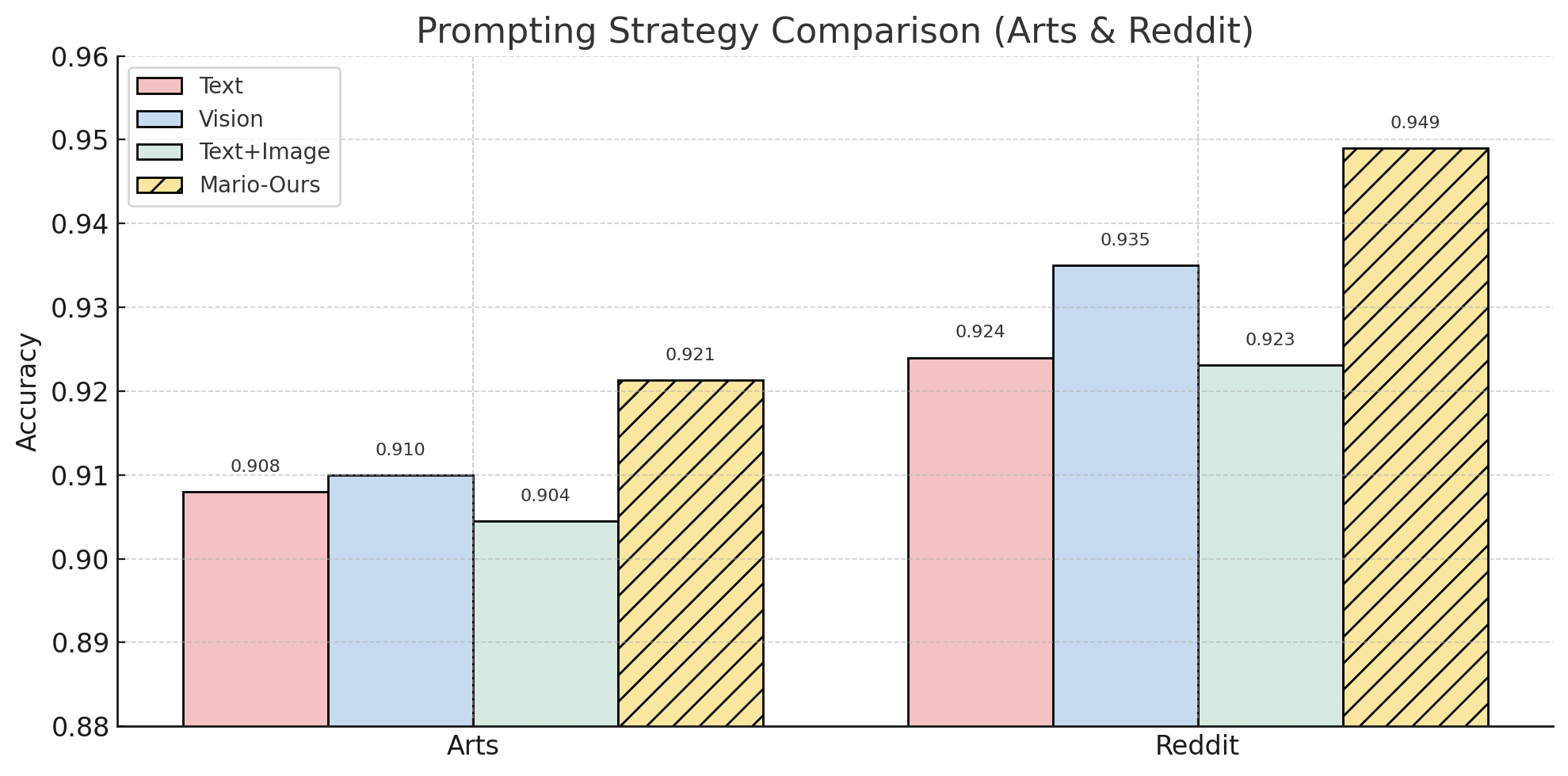}
        \caption{Arts\&Reddit}
        \label{fig:3b}
    \end{subfigure}
    \hfill
    \begin{subfigure}[b]{0.48\linewidth}
        \includegraphics[width=\linewidth]{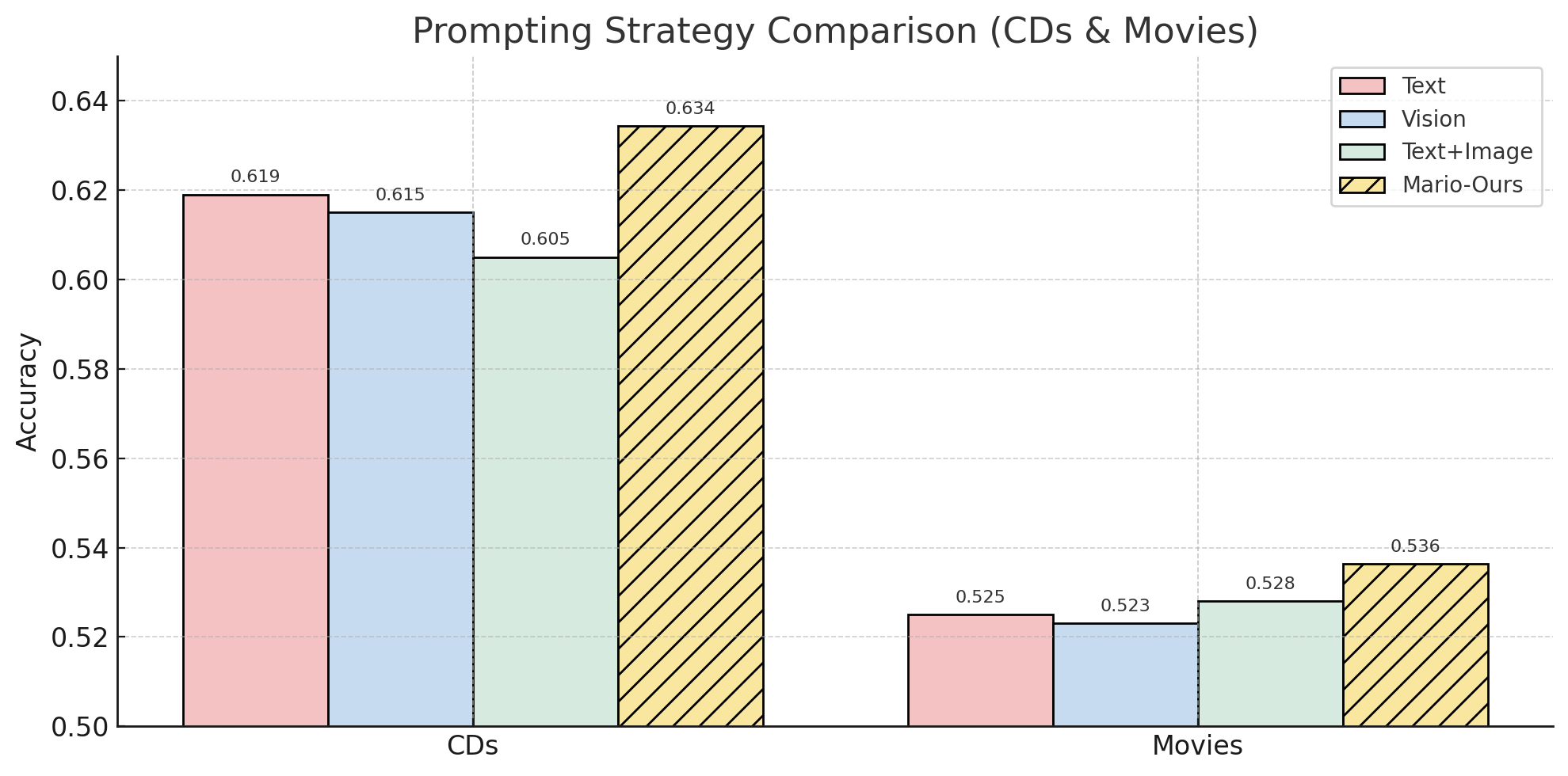}
        \caption{CDs\&Movies}
        \label{fig:3c}
    \end{subfigure}
    \vspace{-0.3cm}
    \caption{Comparison of Mario with three fixed prompt templates containing different modality information across the four datasets.}
    \vspace{-0.6cm}
    \label{fig:3bc}
\end{figure}

\textit{\textbf{\underline{Observation 6:} With Modality-Adaptive Graph Instruction Tuning, LLMs exhibit notably faster convergence and consistently outperform all single-template counterparts by a large margin.}} Fig.~\ref{fig:3a} compares the training losses of Mario against a single-template variant (w/o MAPR). As shown, Mario achieves significantly faster convergence on both Movies \textbf{(2.3$\times$)} and Reddit \textbf{(1.3$\times$)}, while also attaining lower final losses after convergence. Although each epoch of Mario takes approximately 1.5-2$\times$ longer than the single-template variant observed in our experiments, its accelerated convergence enables it to complete training in a comparable overall time.
Fig.~\ref{fig:3bc}  provides a detailed comparison between Mario and different single-template variants. Benefiting from its adaptive tuning mechanism, Mario consistently outperforms all variants by a large margin—for example, on the CDs dataset, it relatively surpasses the average performance of variants by \textbf{3.4\%}. This indirectly demonstrates the importance of respecting nodes' modality preferences.

\begin{figure}[ht]
    \centering
    \begin{subfigure}[b]{0.48\linewidth}
        \includegraphics[width=\linewidth]{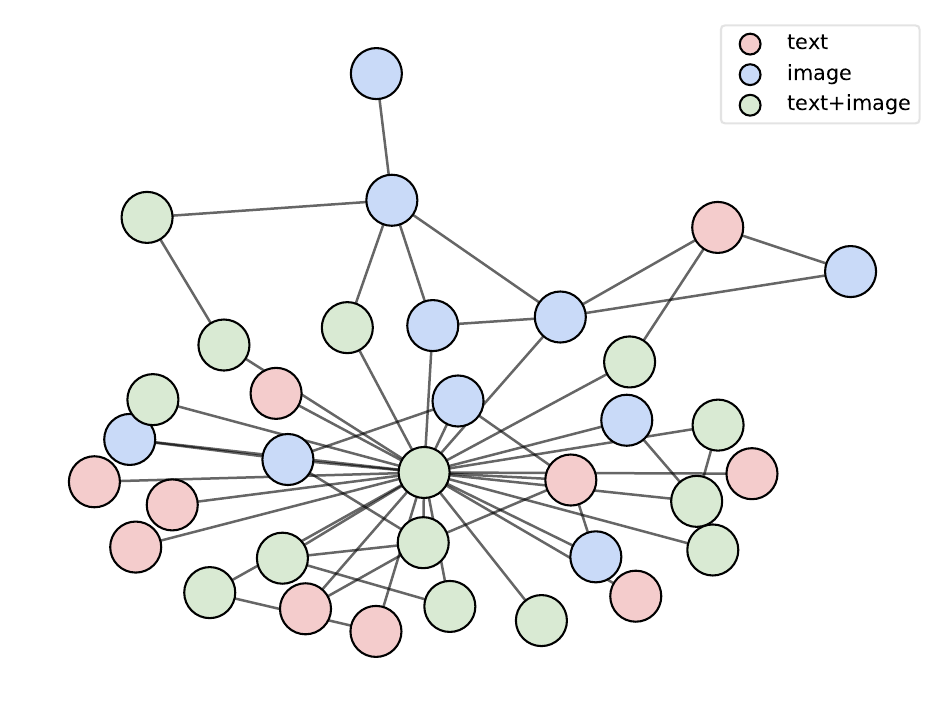}
        \caption{Movies}
        \label{fig:3d}
    \end{subfigure}
    \hfill
    \begin{subfigure}[b]{0.48\linewidth}
        \includegraphics[width=\linewidth]{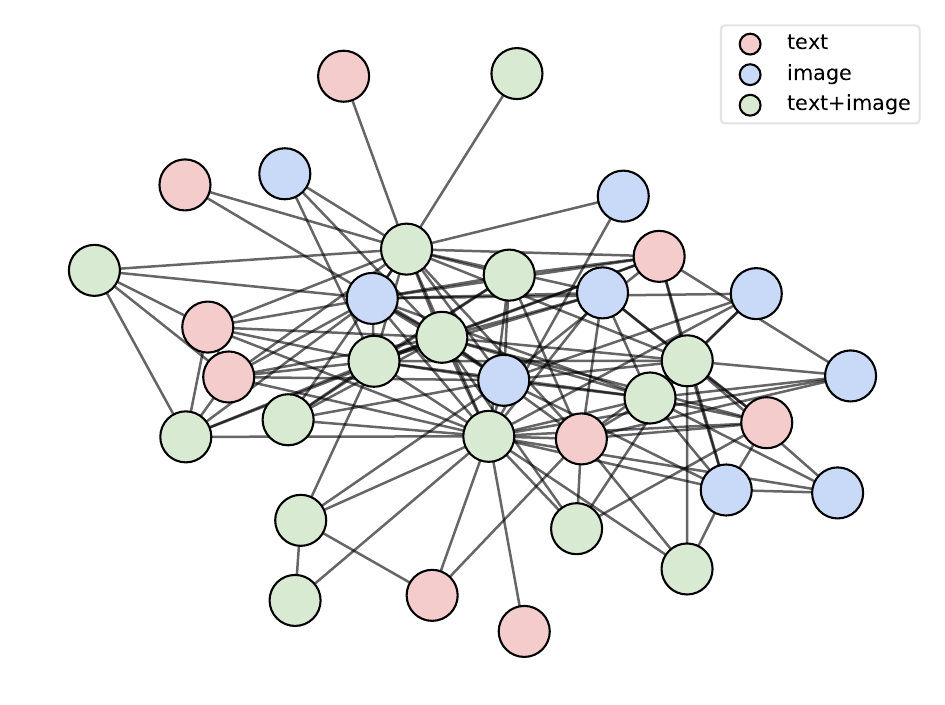}
        \caption{Arts}
        \label{fig:3e}
    \end{subfigure}
    \vspace{-0.2cm}
    \caption{Visualization of Router Selections across two MMGs.}
    \vspace{-0.4cm}
    \label{fig:3de}
\end{figure}

\textit{\textbf{\underline{Observation 7:} Modality preferences in MMGs largely follow a homophily pattern.}}
To make this concrete, Fig.~\ref{fig:3de} visualizes MAPR's modality choice for each node within two larger subregions sampled from two MMGs.
Within each region, the distribution of modality preferences is clearly non-uniform: nodes with the same color (preferred modality) often appear in small clusters, and certain areas are dominated by a single modality (e.g., green text+image nodes in Arts in the middle).
This suggests that neighboring nodes, which are connected because users tend to co-view or co-purchase the corresponding items, often share similar semantic attributes and thus benefit from similar “best” modalities.
We observe analogous locally coherent patterns in many other parts of the above two graphs as well.

\section{Conclusion}
In this paper, we highlight two underexplored challenges in MMG reasoning: cross-modal inconsistency and heterogeneous modality preference. We propose Mario, a novel unified two-stage framework that performs structure-aware image–text alignment with a graph-conditioned vision–language model, then applies modality-adaptive graph instruction tuning via a lightweight router that learns node-specific routing to satisfy the nodes' modality preferences. Extensive experiments on multiple MMG benchmarks show that Mario consistently outperforms strong baselines and enables more reliable multimodal graph reasoning. We hope our work paves the way for future advances in LLM-based multimodal graph reasoning.

{
    \small
    \bibliographystyle{ieeenat_fullname}
    \bibliography{main}
}

\maketitlesupplementary
\section{Appendix}
\subsection{Dataset Details}

\noindent\textbf{Statistics and Introduction.} The detailed statistics of the datasets we used are shown in Table~\ref{tab:statistics}. In these datasets, nodes represent individual products or posts, and edges denote relationships such as co-purchase or co-comment interactions between products or posts. Each node is assigned a label corresponding to its category. Every node is enriched with two modalities: textual attributes, such as product titles, descriptions, or post content, and visual attributes extracted from associated product or post images. Unlike conventional image–text benchmarks where captions are written to explicitly describe the visual content, here the two modalities are only loosely coupled and often contain complementary or even disjoint information. For example, a clothing item may have a textual description that focuses on material and fit (e.g., “soft cotton hoodie with relaxed, oversized silhouette, ideal for fall weather”) while its image emphasizes color, style, and brand logos that are never mentioned in the text. Conversely, the text may include attributes such as size range, discount information, or user-targeted marketing slogans that are not visually observable.

\noindent\textbf{Data Splits.} For the node classification task, we adopt a standardized 6:2:2 split into training, validation, and testing for Mario and all the baselines. For the link prediction task, the training, validation, and test sets consist of 3,000, 2,000, and 1,000 edges, respectively, for training and evaluation. 

\begin{table}[H]
\centering
\small
\setlength{\tabcolsep}{3pt}
\caption{Dataset statistics across multiple MMG datasets.}
\label{tab:statistics}
\begin{tabular}{ccccc}
\toprule
\textbf{Dataset} & \textbf{Domain} & \textbf{\# Nodes} & \textbf{\# Edges} & \textbf{\# Classes} \\
\midrule
Movies     & E-commerce & 16,672   & 109,195    & 20 \\
Toys       & E-commerce    & 20,694   & 63,443     & 18 \\
CDs        & E-commerce    & 36,272   & 844,878    & 15 \\
Arts       & E-commerce    & 28,195   & 197,428    & 7  \\
Reddit(S)  & Social Media  & 15,894  & 566,160 & 20 \\
Goodreads  & Literature    & 685,294  & 7,235,084  & 11  \\
\bottomrule
\end{tabular}
\end{table}

\subsection{Experiment Details}
\label{experiments_appendix}
In this section, we provide additional explanations for  experiment details not covered in the paper.

\noindent\textbf{Image to caption conversion.}
Since current GraphLLM baselines do not support processing image features, we convert images into captions using VLMs in the text+vision experiments to enhance textual modality with auxiliary information. This facilitates multimodal graph reasoning. The model used for this purpose is Qwen-VL-Chat \cite{bai2023qwen}.

\noindent\textbf{L(V)LMs-Based Baseline Experiment Execution.}
In the experiments, we frequently mention using LLaMA and LLaVA. All these experiments were conducted with the assistance of vLLM \cite{kwon2023efficient}. vLLM is a high-performance library for efficient LLM inference and serving. It provides state-of-the-art serving throughput with optimizations such as PagedAttention, continuous batching, CUDA acceleration, FlashAttention, and speculative decoding, ensuring low-latency execution. vLLM seamlessly integrates with Hugging Face models, supports various decoding strategies, and enables tensor/pipeline parallelism across diverse hardware platforms. In our experiments, we utilized vLLM to efficiently serve LLaMA and LLaVA, enabling scalable inference for text-based and MMG reasoning tasks while ensuring computational efficiency and high throughput.

\noindent\textbf{Hyper-Parameter Settings.}
We provide a detailed discussion of the hyper-parameter settings used in our experiments. For Stage 1, we usually employ one layer (up to two) of GraphTransformer for structure-aware text-image alignment and we sample \textbf{$\sim$10} nodes ($\mathcal V_s$) to feed into the GVLM. For Stage 2, we typically select 10-15 neighbors to provide neighbor context and conduct 10 epochs of instruction tuning using LLaMA3.1-8B with early stop strategy. We use a four-layer MLP as the MAPR, and set $\lambda$ = 0.01. For link prediction experiments, we only provide the neighbor context of the first node in the prompt, but these are common neighbors with the other node. The projection layer consists of two layers. For GraphPrompter \cite{liu2024can}, we use LLaMA3.1-8B as the final LLM for inference. For LLaGA \cite{chenllaga}, we follow the original paper and adopt the same setting, where Vicuna \cite{chiang2023vicuna} serves as the primary foundational large language model. We truncate the final tokens input length to 512. All experiments involving LLM deployment were conducted on two A100-SXM4-80-GB GPUs. For GraphLLM-based baselines, we did not evaluate the vision-only setting. This is because such frameworks are inherently text-centric by design, and we followed their original modeling philosophy without extending them to vision-only scenarios. Additionally, we experimented with using image captions alone to support inference within these models, but the performance was significantly worse compared to text-only or image+text settings. Therefore, we did not include the vision-only results in the paper.

\begin{figure*}[t]
    \centering
    \begin{subfigure}[b]{0.3\textwidth}
        \centering
        \includegraphics[width=\linewidth]{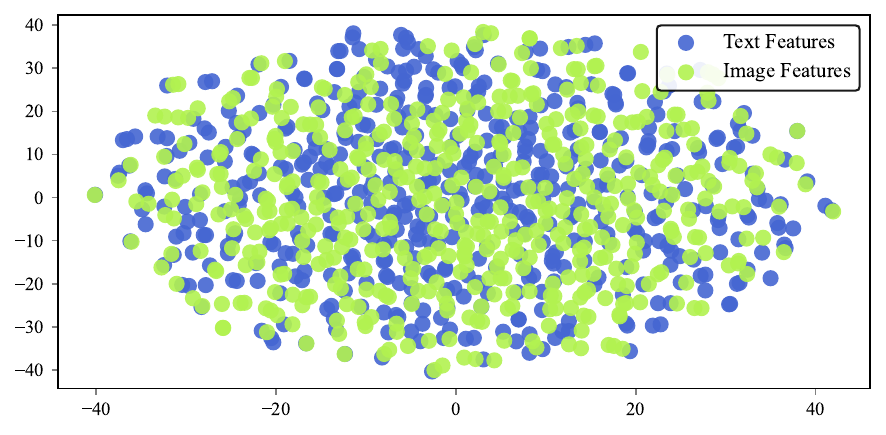}
        \caption{Movies – Frozen CLIP}
        \label{fig:tsne_movies_model1}
    \end{subfigure}
    \hfill
    \begin{subfigure}[b]{0.3\textwidth}
        \centering
        \includegraphics[width=\linewidth]{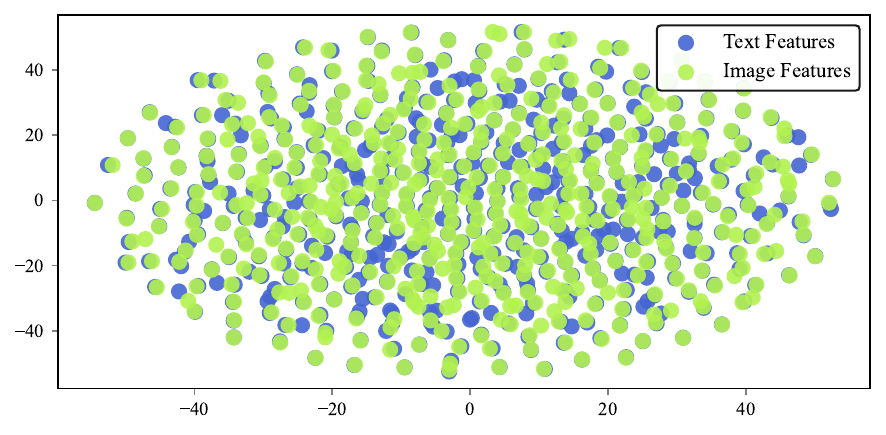}
        \caption{Movies – Tuned CLIP}
        \label{fig:tsne_movies_model2}
    \end{subfigure}
    \hfill
    \begin{subfigure}[b]{0.3\textwidth}
        \centering
        \includegraphics[width=\linewidth]{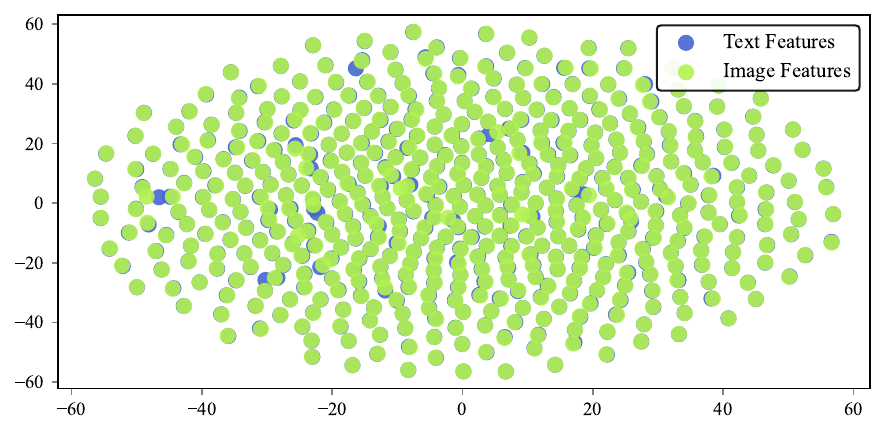}
        \caption{Movies – Mario's GVLM (ours)}
        \label{fig:tsne_movies_mario}
    \end{subfigure}

    \vspace{0.6em}

    \begin{subfigure}[b]{0.3\textwidth}
        \centering
        \includegraphics[width=\linewidth]{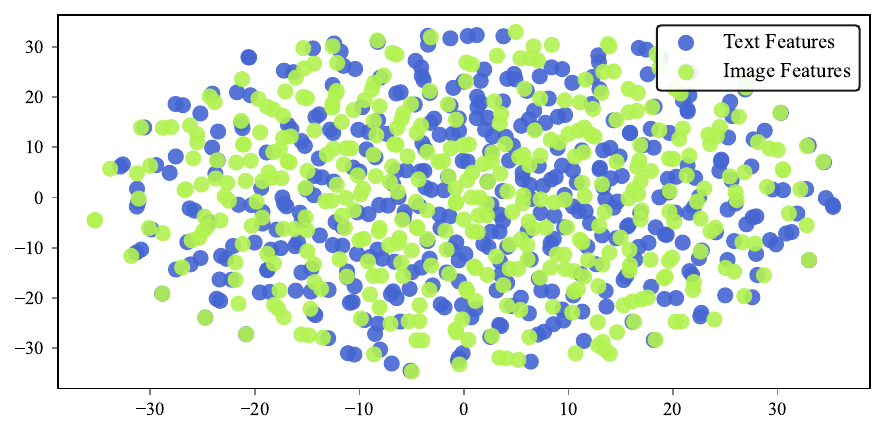}
        \caption{Reddit – Frozen CLIP}
        \label{fig:tsne_reddit_model1}
    \end{subfigure}
    \hfill
    \begin{subfigure}[b]{0.3\textwidth}
        \centering
        \includegraphics[width=\linewidth]{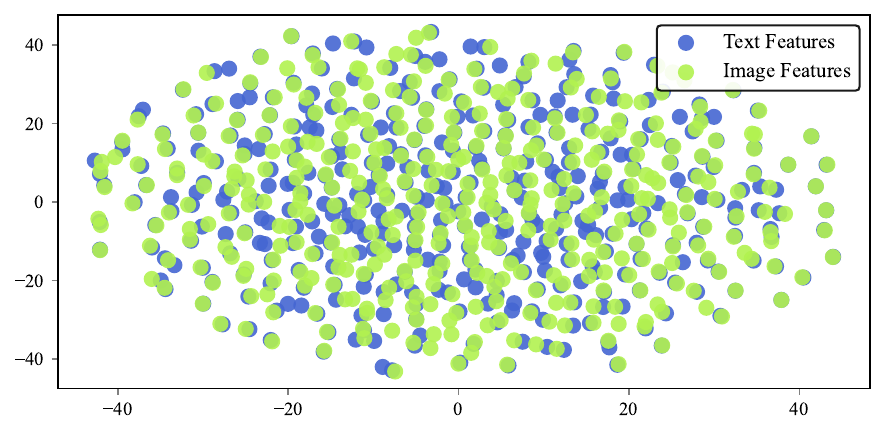}
        \caption{Reddit – Tuned CLIP}
        \label{fig:tsne_reddit_model2}
    \end{subfigure}
    \hfill
    \begin{subfigure}[b]{0.3\textwidth}
        \centering
        \includegraphics[width=\linewidth]{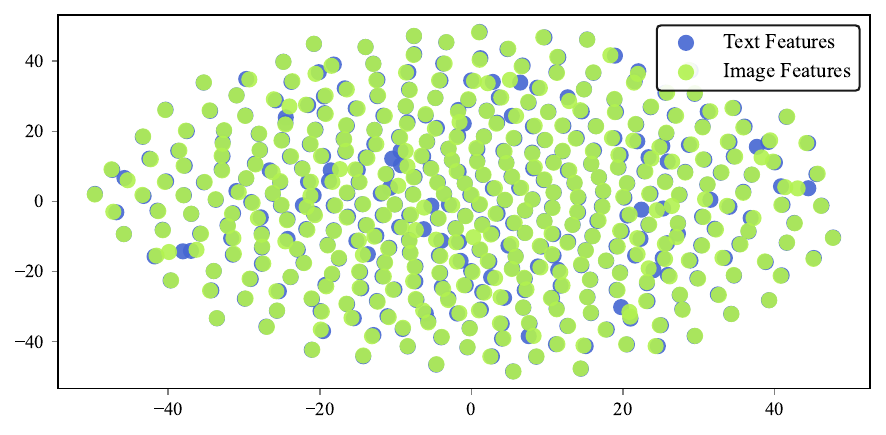}
        \caption{Reddit – Mario's GVLM (ours)}
        \label{fig:tsne_reddit_mario}
    \end{subfigure}

    \caption{
    t-SNE visualizations of aligned multimodal features on \textit{Movies} (top) 
    and \textit{Reddit} (bottom) for the three models in Fig.~1. 
    For each dataset, we project a randomly sampled subset of nodes from the full graph, 
    using their aligned text and image representations as input to t-SNE. 
    Comparing the six panels reveals how different alignment strategies affect 
    the relative organization of text and image features in the shared latent space.
    }
    \label{fig:tsne_alignment}
\end{figure*}

\subsection{t-SNE Visualization of GVLM Alignment}
\label{sec:tsne_appendix}

To further illustrate the qualitative differences between the three models in Fig.~1, 
we visualize their aligned text and image features using t-SNE on two multimodal graphs, 
\textit{Movies} and \textit{Reddit}. For each dataset, we randomly sample a subset of nodes 
from the full graph and project their aligned text/image representations to 2D. 
This subsampling allows us to focus more closely on the structural differences between models 
while still capturing representative patterns. The six panels in Fig.~\ref{fig:tsne_alignment} 
show the resulting distributions for the three models on \textit{Movies} (top row) and 
\textit{Reddit} (bottom row), respectively. Beyond the overall layout, we observe consistent qualitative trends across the six panels in
Fig. 1 in the paper. On both \textit{Movies} and \textit{Reddit}, the frozen CLIP
features form two loosely overlapping clouds, indicating a sizeable gap between text and image
representations. Fine-tuning CLIP shrinks this gap and slightly tightens the clusters, but the
two modalities still remain partially misaligned. In contrast, Mario produces a much more
intertwined manifold where text and image features are almost co-located and organized along
smoother global structures, suggesting that our graph-conditioned alignment achieved by Mario's GVLM substantially
improves cross-modal consistency while preserving meaningful semantic variation.

\subsection{Comparison with MMGCN and MGAT}
\label{sec:mmgcn_mgat_appendix}

In the main paper, MMGCN \cite{wei2019mmgcn} and MGAT \cite{tao2020mgat} are excluded, as they focus primarily on
recommendation-style tasks and showed weak performance in our setting through
initial experiments. For completeness, we provide here a small-scale comparison
to substantiate this design choice. Table~\ref{tab:mmgcn_mgat_nc} reports their
node classification accuracy on \textit{Movies} and \textit{Arts}, alongside
representative "text+image" GNN baselines under the same
experimental protocol.

\begin{table}[t]
    \centering
    \small
    \setlength{\tabcolsep}{16pt}
    \renewcommand{\arraystretch}{1.0}
    \caption{Node classification accuracy (\%) on \textit{Movies} and \textit{Arts} for additional multimodal baselines (MMGCN, MGAT) and representative unimodal GNNs (text+image settings).}
    \label{tab:mmgcn_mgat_nc}
    \begin{tabular}{lcc}
        \toprule
        \textbf{Model} & \textbf{Movies} & \textbf{Arts} \\
        \midrule
        SAGE   & 44.07 & 85.35 \\
        GATv2  & 49.29 & 81.19 \\
        GCN  & 46.96 & 76.76 \\
        MMGCN  & 46.79 & 86.63 \\
        MGAT   & 40.17 & 87.25 \\
        \bottomrule
    \end{tabular}
\end{table}

Overall, MMGCN and MGAT do not show clear advantages over standard GNNs.
On \textit{Movies}, MMGCN is essentially on par with GCN and still below
GATv2, while MGAT performs worse than all three GNN baselines. On
\textit{Arts}, MMGCN and MGAT slightly outperform some GNNs, but the gains are
modest and all these methods remain far from the strong multimodal models and
Mario reported in the main tables. Since SAGE, GATv2, and GCN are already
treated as weak baselines in our core comparison, adding MMGCN and MGAT there
would not change the conclusions; we therefore only include them in this
appendix section for completeness. A similar conclusion can also be drawn from MLaGA \cite{fan2025mlaga}.

\begin{table*}[ht]
\centering
\caption{Frozen Mario versus LoRA-Tuned Mario (Accuracy \%).}
\begin{tabular}{c|c|c|c|c|c}
\toprule
\textbf{Model} & \textbf{Trainable Params} & \textbf{Movies} & \textbf{Reddit} & \textbf{CDs} & \textbf{Arts} 
\\ \hline
\multicolumn{6}{c}{\textbf{Node Classification (Trainable parameters are from Stage 2)}} 
\\ \hline
Frozen Mario       
& 18,886,656 (0.2346\%)  & 50.85  & 93.60  & 60.45   & 89.69                  \\ \hline
Mario + LoRA         
& 22,294,528 (0.2768\%) & 53.63  & 95.30 & 63.43   & 92.13      \\ \hline
\multicolumn{6}{c}{\textbf{Link Prediction (Trainable parameters are from Stage 2)}} 
\\ \hline
Frozen Mario         
& 18,886,656 (0.2346\%)  & 90.90  & 89.00   & 88.60   & 86.30                  \\ \hline
Mario + LoRA      
& 22,294,528 (0.2768\%) & 93.90  & 91.30  & 92.70 & 89.96      \\ \bottomrule
\end{tabular}
\label{Frozen Mario}
\end{table*}

\subsection{Additional GNNs Zero-Shot Results}
In our zero-shot experiments in the paper, we assess the transferability of graph neural networks (GNNs) to new datasets, without re-training their core parameters. Specifically, when transitioning between datasets, we retain the trained GNN model, including its network architecture and learned parameters, and only replace the classifier layer corresponding to the new dataset. This approach ensures that the underlying graph feature extractor remains unchanged, allowing us to evaluate the generalization capacity of different models under domain shifts.

Table ~\ref{tab:transfer_appendix} presents the additional zero-shot transfer results across different models. This result serves as a supplement to Table 3 in the paper (where GraphLLMs adopt the text-only setting, and the other baselines adopt the text+vision setting). For the results below without explicit modality specification, the \textbf{text-only modality} is used (different from the setting in the paper). We evaluate the same two transfer settings: (1) Toys → Movies, where models trained on the Toys dataset are directly applied to the Movies dataset, and (2) Toys+Movies → CDs, where models trained on both the Toys and Movies datasets are tested on the CDs dataset. The evaluation is conducted under two tasks: NC (Node Classification Accuracy) and LP (Link Prediction Accuracy).

Across both transfer settings, our Mario significantly outperforms all baselines, demonstrating strong zero-shot adaptation capabilities. In contrast, traditional GNNs such as GCN, GATv2, and SAGE struggle to generalize, exhibiting considerably lower performance. For instance, in the Toys → Movies setting, GCN achieves an NC score of only 3.29, while Mario achieves 41.00, more than 10 times higher. A similar trend is observed in Toys+Movies → CDs, where Mario attains an NC score of 54.32, substantially outperforming all baselines.

Furthermore, while MLP-based models (both text-only and vision-only versions) show moderate performance in link prediction, they underperform in node classification due to their inability to leverage structural dependencies effectively. These results underscore the limitations of conventional GNNs in zero-shot scenarios and highlight the advantages of our Mario model in learning transferable multimodal representations.

\begin{table}[ht]
\centering
\small
\setlength{\tabcolsep}{5pt}
\renewcommand{\arraystretch}{1.2}
\caption{Zero-Shot Results (Accuracy \%). }
\label{tab:transfer_appendix}
\begin{tabular}{c|cc|cc}
\toprule
\multirow{2}{*}{\textbf{Model}} & \multicolumn{2}{c}{\textbf{Toys $\rightarrow$ Movies}} & \multicolumn{2}{|c}{\textbf{Toys+Movies $\rightarrow$ CDs}} \\
\cline{2-5}
 & \textbf{NC} & \textbf{LP} & \textbf{NC} & \textbf{LP} \\
\midrule
MLP             & 6.12 & 52.60 & 7.04 & 50.20  \\
GCN             & 3.29 & 62.13 & 10.01 & 64.17  \\
GATv2           & 4.32 & 64.47 & 8.13 & 67.97  \\
SAGE            & 3.11 & 55.83 & 6.14 & 59.63 \\
MLP(Vision Only)& 4.61 & 52.13 & 9.06 & 46.09 \\
\hline
Mario-8B (Ours) & \textbf{41.00} & \textbf{86.60} & \textbf{54.32} & \textbf{82.50}  \\
\bottomrule
\end{tabular}
\end{table}

\subsection{Frozen vs. LoRA-Tuned Mario}
We also find that LoRA-tuned Mario outperforms its frozen counterpart, and \textit{\textbf{both exceed all baselines by a large margin}}.
As shown in Table~\ref{Frozen Mario}, LoRA tuning yields consistent gains of about 1.7–3.0 points in node classification accuracy across all four datasets (e.g., from 50.85 to 53.63 on \textit{Movies} and from 89.69 to 92.13 on \textit{Arts}), and similarly improves link prediction by roughly 2–4 points.
These improvements come with only a tiny increase in the number of trainable parameters, from 18.9M (0.2346\%) to 22.3M (0.2768\%) of the full LLM, indicating that Mario is already strong in a frozen-LLM regime while a lightweight LoRA adapter can further boost performance without sacrificing parameter efficiency.

\begin{figure*}
    \centering
    \includegraphics[width=0.8\linewidth]{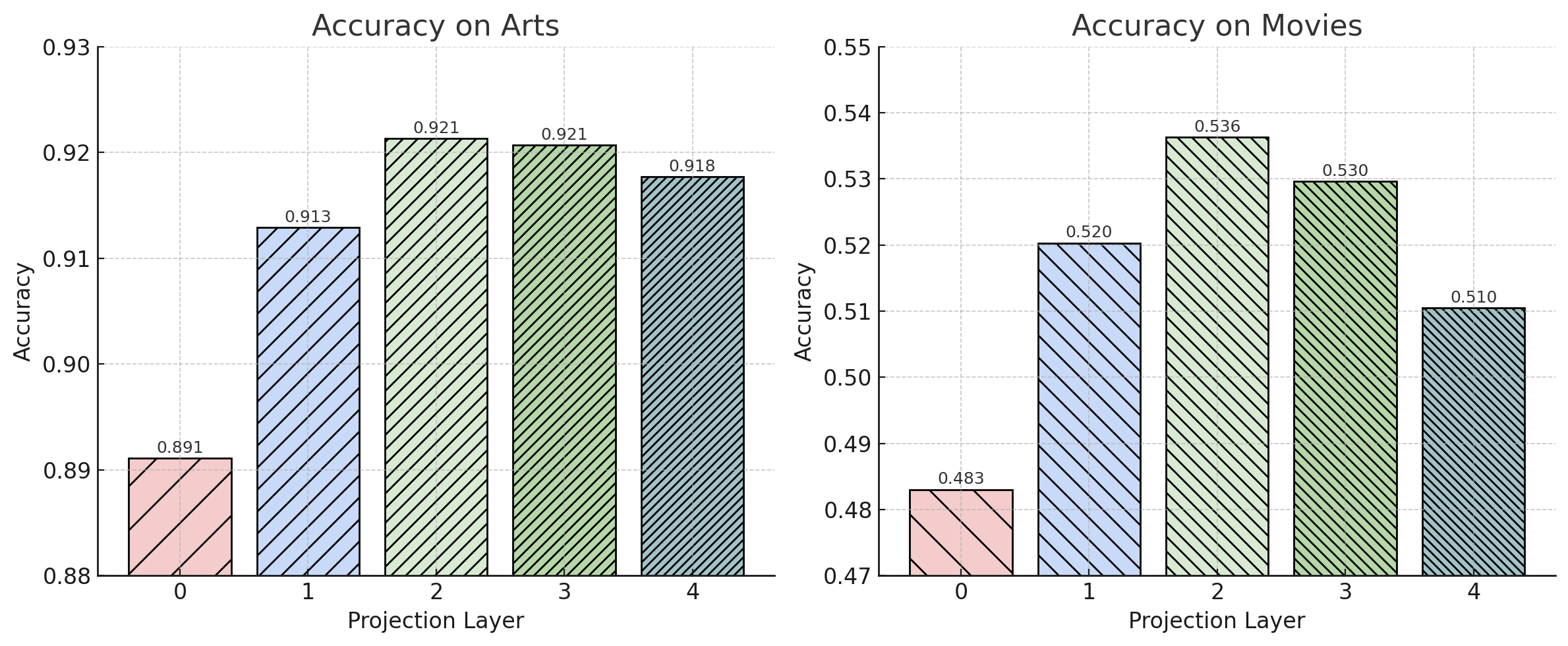}
    \caption{Sensitivity analysis of the projection layer in Arts and Movies}
    \label{fig:appendixf1}
\end{figure*}

\begin{figure*}
    \centering
    \includegraphics[width=0.8\linewidth]{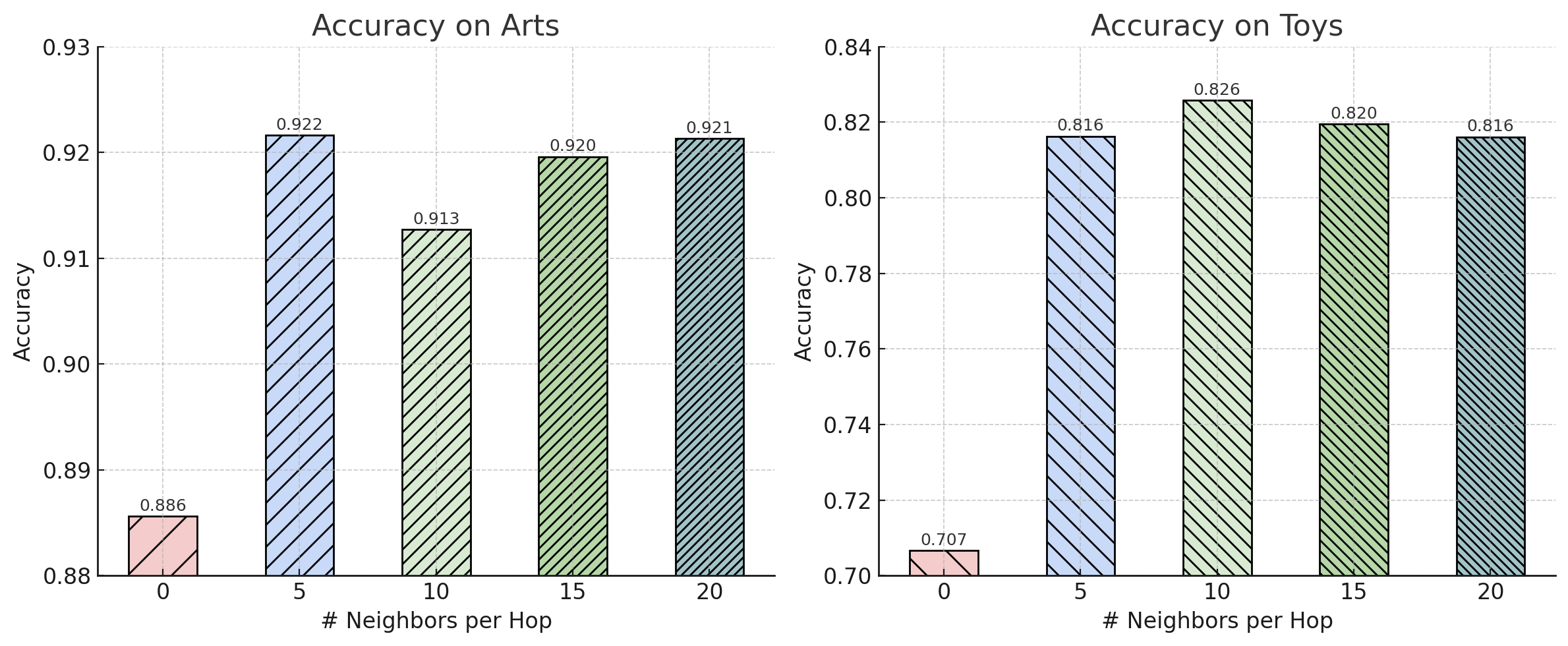}
    \caption{Sensitivity analysis of the number of neighbors per hop in Arts and Toys}
    \label{fig:appendixf2}
\end{figure*}

\subsection{Ablation Study of LLM Backbone}
To assess the robustness of Mario across different LLMs, we conduct an ablation study using a range of LLM backbones, including both LLaMA-based and non-LLaMA families. As summarized in Table~\ref{tab:mario_sizes}, Mario consistently delivers strong performance regardless of the specific LLM used, highlighting the generalizability of our framework.

Within the LLaMA2 family, increasing model size from 7B to 13B results in negligible improvement: on Arts, accuracy rises slightly from 91.06\% to 91.23\%, while performance on Toys slightly drops from 81.20\% to 80.93\%. Similarly, when switching from LLaMA2 to Vicuna-v1.5 (also LLaMA2-based), results remain largely consistent, indicating that mere scaling or minor tuning of the base LLM does not significantly alter performance in our multimodal graph reasoning tasks.

More importantly, Mario remains effective even when paired with structurally different LLMs. Using FLAN-T5-XXL, a T5-style encoder-decoder model, Mario achieves 92.08\% on Arts and 81.63\% on Toys, outperforming all LLaMA2 variants. Furthermore, Mario-8B (LLaMA3), our best-performing configuration, achieves 92.13\% and 82.58\% on Arts and Toys respectively, demonstrating stronger capability than its LLaMA2 predecessors.

These observations collectively indicate that Mario's architectural design—rather than the choice of LLM backbone—is the key contributor to its strong performance. Whether applied to decoder-only (LLaMA), instruction-tuned (Vicuna), or encoder-decoder (FLAN-T5) models, Mario exhibits consistent gains, underscoring its backbone-agnostic robustness in multimodal graph reasoning.

\begin{table}[htbp]
    \centering
    \caption{Ablation Study of Different LLMs. (Accuracy \%)}
    \label{tab:mario_sizes}
    \begin{tabular}{c|c|c}
        \toprule
        \textbf{Different Size} & \textbf{Arts} & \textbf{Toys} \\
        \midrule
        Mario-7B (LLaMA2)  & 91.06  & 81.20  \\
        Mario-13B (LLaMA2) & 91.23  & 80.93  \\
        Mario-7B (Vicuna-v1.5) & 91.09  & 81.07  \\
        Mario (FLAN-T5-XXL) & 92.08  & 81.63  \\
        \textbf{Mario-8B (LLaMA3)}  & \textbf{92.13} & \textbf{82.58} \\
        \bottomrule
    \end{tabular}
\end{table}

\begin{table*}[htbp]
\centering
\caption{Heterophily Ratios of Benchmark Datasets}
\label{tab:heterophily}
\begin{tabular}{lcccccc}
\toprule
\textbf{Dataset} & Movies & Toys & Arts & CDs & Goodreads & Reddit \\
\midrule
\textbf{Heterophily Ratio} & 0.53 & 0.26 & 0.34 & 0.69 & 0.33 & 0.04 \\
\bottomrule
\end{tabular}
\end{table*}

\subsection{Sensitivity Analysis}

To assess the effectiveness of our designed instruction templates that incorporate multimodal node features, we conducted a sensitivity analysis on two critical components: the number of projection layers and the length of the neighbor context. These factors directly influence how effectively multimodal information is aligned and delivered to the LLM for reasoning.
As shown in Figure~\ref{fig:appendixf1}, introducing a projection layer consistently improves performance over the baseline without projection. Notably, employing two layers yields the best or near-best results across both Arts and Movies datasets. This suggests that a lightweight projection module facilitates better multimodal alignment without incurring excessive complexity, enhancing the model's ability to interpret visual-textual signals.

Similarly, Figure~\ref{fig:appendixf2} illustrates that incorporating a limited number of neighbors per hop significantly boosts performance compared to the zero-neighbor setting. For instance, in the Toys dataset, adding neighbor context improves accuracy by over 10\%. However, further increasing the number of neighbors yields marginal or unstable gains, indicating that a moderate amount of structural context is optimal. These results highlight the importance of integrating a controlled amount of structural information into the prompt, allowing the LLM to better contextualize the target node during reasoning.

\subsection{Variance Analysis}

Following prior GraphLLM studies, we initially omitted variance reporting. However, our experiments reveal that the variance of our method is relatively small—typically around \(\pm 0.07\) or \(\pm 0.14\) across three random runs. For reference, Table~\ref{tab:variance_results} summarizes partial variance scores on representative datasets and tasks (Metric: Accuracy).

\begin{table}[htbp]
    \centering
    \caption{Partial variance results of Mario across datasets and tasks over 3 runs.}
    \label{tab:variance_results}
    \begin{tabular}{lcc}
        \toprule
        \textbf{Method} & \textbf{Movies (NC)} & \textbf{Arts (LP)} \\
        \midrule
        Mario (Single Focus) & $53.63 \pm 0.07$ & $89.96 \pm 0.14$ \\
        Mario (Mix Training) & $50.98 \pm 0.08$ & $92.60 \pm 0.12$ \\
        \bottomrule
    \end{tabular}
\end{table}

\subsection{Quantitative Analysis of Modality Preference}
\label{sec:7.10}
This subsection explains the statistics in the Venn diagram of Figure~\ref{Fig1}. Specifically, the six numbers in Figure~\ref{Fig1}(b) can be grouped into three categories: (i) the proportion of nodes that can be correctly classified \emph{only} by the template corresponding to a single modality; (ii) the proportion of nodes that can be correctly classified \emph{only} when two templates are both correct (rather than counting a node as correct if either template is correct); and (iii) the proportion of nodes that can be correctly classified by all three template types. The proportions in the first category are 2.65\%, 2.25\%, and 2.05\%; those in the second category are 7.71\%, 7.40\%, and 6.98\%; and the third category accounts for 70.96\%. These numbers sum to 100\%. Therefore, the statement that ``about 30\% of nodes cannot be correctly classified by all templates jointly'' is computed as $100\% - 70.96\% = 29.04\%$, which is approximately 30\%. All percentages are normalized within the set of nodes correctly classified by at least one template.

\subsection{Robustness against Varying Heterophily}
To reduce overfitting to locally uniform neighborhoods and to expose the model to richer semantic context, we adopt a multi-hop neighbor selection strategy. Expanding the receptive field beyond immediate neighbors allows Mario to retrieve distant yet relevant nodes, so the router is not forced to rely solely on short-range label similarity when forming prompts.
To quantify structural diversity in our benchmarks, we compute each graph’s heterophily ratio, defined as the fraction of edges linking nodes with different labels.

Despite the heterophily ratios varying widely across datasets—from near-homophilic graphs such as Reddit (0.04) to strongly heterophilic ones like CDs (0.69) and Movies (0.53)—Mario consistently maintains strong performance. This suggests that our Stage-1 feature–based similar neighbor selection remains reliable across different structural regimes, confirming that it generalizes well even when local neighborhoods are not label-coherent. Importantly, while this finding is complementary to Observation 7 in the paper, it addresses a different question: here we show robustness of the selection strategy under varying heterophily, rather than characterizing the spatial pattern of modality preferences within the graph.

\subsection{Training Compute Analysis}
We compare Mario’s Stage 1/2 with all baselines on identical data under a compute-matched budget (Table~\ref{tab:trainingcompute}), which reports the training cost and resulting performance on our main datasets. To ensure that every method had sufficient opportunity to converge, we initially capped each Stage-1 run at 2 GPU-hours on A100 SXM4 80GB GPUs and terminated runs that remained unconverged at the cap. In practice, however, we observed that the GVLM and all graph-based baselines consistently converged within 1 GPU-hour, with no notable gap in training overhead across methods, indicating that our comparisons are conducted under a largely fair and compute-balanced setting. We further include a new Tuned CLIP baseline for a stronger reference; since its optimization is typically more compute-intensive than GVLM/GNN-style training, we set its compute cap to 1 GPU-hour as well to maintain fairness given that the other methods already converge within this budget. As shown in Table~\ref{tab:trainingcompute} (the lower part), MAPR converges
in roughly half the epochs of the baselines (stage2), with an average 
total runtime only 0.25 (rather than 3×) GPU-hours higher. Finally, the resulting average accuracies under these compute caps closely match those reported in Table~\ref{tab:stage1_results} and Figure~\ref{fig:3bc}, with no noticeable discrepancies.

\begin{table}[htbp]
    \centering
    \caption{Detailed Cost Breakdown  (Stage 1 \& 2).}
    \scriptsize
    \setlength{\tabcolsep}{2.9pt}
    \renewcommand{\arraystretch}{1.5}
    \begin{tabular}{c||c|c|c|c|c}
        \hline
        \rowcolor{CadetBlue!20}
        \textbf{Method} & \textbf{Arts} & \textbf{Reddit} & \textbf{Movies} & \textbf{\#Epoch} & \textbf{Avg Acc(\%)} \\
        \hline

        \multicolumn{6}{c}{\textbf{Tuned CLIP/Other Baselines vs. GVLM (stage1) (Columns 2–4: GPU-hours)}} \\
        \hline

        GCN
        & 0.95
        & 0.94
        & 0.88
        & 36.9
        & 78.17 \myred{2.79}\\

        SAGE
        & 0.91 
        & 0.90 
        & 0.88
        & 34.3
        & 77.98 \myred{3.04}\\

        GATv2
        & 0.93
        & 0.91
        & 0.89
        & 33.0
        & 78.08 \myred{2.92}\\

        MLP
        & 0.90
        & 0.87
        & 0.85
        & 37.0
        & 77.78 \myred{3.30}\\ \hline

        Tuned CLIP
        & 0.99
        & 0.98
        & 0.99
        & 22.3
        & 78.01 \myred{3.00}\\

        \hline
        \rowcolor[HTML]{D7F6FF}
        \textbf{GVLM}
        & 0.92
        & 0.91
        & 0.88
        & 23.0
        &  80.35 \\
        \hline

        \multicolumn{6}{c}{\textbf{Single-template Variants vs. MAPR (Stage2) (Columns 2–4: GPU-hours)}} \\
        \hline

        Text-only
        & 5.65
        & 4.02
        & 4.11
        & 6.3
        & $78.56$ \myred{2.28} \\

        Image-only
        & 5.65
        & 4.58
        & 4.11
        & 6.6
        & $78.93$ \myred{1.80} \\

        Text+Image
        & 5.77
        & 4.09
        & 4.23
        & 6.3
        & $78.50$ \myred{2.36} \\

        \hline
        \rowcolor[HTML]{D7F6FF}
        \textbf{Mario (MAPR)}
        & 5.82
        & 4.15
        & 4.35
        & 3.0
        & \textbf{$80.35$} \\
        \hline
    \end{tabular}
    \vspace{-0.4cm}   
    \label{tab:trainingcompute}
\end{table}
\subsection{Prompt Template}
The prompt templates used for adaptive multimodal graph instruction tuning in the two multimodal graph reasoning tasks, node classification and link prediction, are shown in Table~\ref{tab:prompt_templates}. Since the templates for different modalities are broadly similar, differing mainly in which modality-specific features of the anchor node and its neighbors are embedded—we present the template for the text+image case as an illustrative example.

\begin{table*}[htbp]
  \centering
  \caption{Prompt Templates for Node Classification and Link Prediction Tasks. Note that this template is designed to include both text and image features of the node. If the input is text-only or image-only, simply retain the corresponding single modality feature.}
  \label{tab:prompt_templates}
  \begin{tabular}{p{0.2\linewidth} p{0.75\linewidth}}
    \toprule
    \textbf{Task} & \textbf{Prompt Template} \\
    \midrule
    Node Classification & 
    \begin{minipage}[t]{\linewidth}
      \small
      I’m starting a node classification task in the \textcolor{blue}{\textless  dataset\textgreater}. Each node represents a \textcolor{blue}{\textless  product\textgreater} with text and image features, and edges indicate \textcolor{blue}{\textless  relationship\textgreater}. Given a target node, the raw text is ...,  the text feature is \textcolor{blue}{\textless text feature\textgreater} and the image feature is \textcolor{blue}{\textless image feature\textgreater}. The neighbors are described in the following template: \textcolor{blue}{\textless text feature\textgreater}, \textcolor{blue}{\textless image feature\textgreater}, and \textcolor{blue}{\textless label\textgreater}.\\[5pt]
      It has the following neighbors at hop 1:\\[3pt]
      N1: \quad \textcolor{blue}{\textless 1-hop neighbor 1 text feature\textgreater}, \textcolor{blue}{\textless 1-hop neighbor 1 image feature\textgreater}, \textcolor{blue}{\textless 1-hop neighbor 1 label\textgreater}\\[3pt]
      N2: \quad \textcolor{blue}{\textless 1-hop neighbor 2 text feature\textgreater}, \textcolor{blue}{\textless 1-hop neighbor 2 image feature\textgreater}, \textcolor{blue}{\textless 1-hop neighbor 2 label\textgreater}\\[3pt]
      N3: ...........\\[5pt]
      It has the following neighbors at hop 2:\\[3pt]
      N1: \quad \textcolor{blue}{\textless 2-hop neighbor 1 text feature\textgreater}, \textcolor{blue}{\textless 2-hop neighbor 1 image feature\textgreater}, \textcolor{blue}{\textless 2-hop neighbor 1 label\textgreater}\\[3pt]
      N2: \quad \textcolor{blue}{\textless 2-hop neighbor 2 text feature\textgreater}, \textcolor{blue}{\textless 2-hop neighbor 2 image feature\textgreater}, \textcolor{blue}{\textless 2-hop neighbor 2 label\textgreater}\\[3pt]
      ......\\[5pt]
      Based on the information provided, please classify the target node into one of the following categories: \textcolor{blue}{\{all\_categories\}}.
    \end{minipage} \\
    \midrule
    Link Prediction & 
    \begin{minipage}[t]{\linewidth}
      \small
      I’m starting a link prediction task in the \textcolor{blue}{\textless dataset\textgreater}. Each node represents a \textcolor{blue}{\textless product\textgreater} with text and image features, and edges indicate \textcolor{blue}{\textless relationship\textgreater}. Given the two nodes:\\[5pt]
      Node 1: The raw text is ... the text feature is \textcolor{blue}{\textless text feature\textgreater}, and the image feature is \textcolor{blue}{\textless image feature\textgreater}.\\[3pt]
      Node 2: The raw text is ... the text feature is \textcolor{blue}{\textless text feature\textgreater}, and the image feature is \textcolor{blue}{\textless image feature\textgreater}.\\[5pt]
      The neighbors of node 1 (common neighbors with node 2) are described in the following template: \textcolor{blue}{\textless text feature\textgreater}, \textcolor{blue}{\textless image feature\textgreater}.\\[5pt]
      It has the following neighbors at hop 1 (Directly connected):\\[3pt]
      N1: \quad \textcolor{blue}{\textless 1-hop neighbor 1 text feature\textgreater}, \textcolor{blue}{\textless 1-hop neighbor 1 image feature\textgreater}\\[3pt]
      N2: \quad \textcolor{blue}{\textless 1-hop neighbor 2 text feature\textgreater}, \textcolor{blue}{\textless 1-hop neighbor 2 image feature\textgreater}\\[3pt]
      N3: ...........\\[5pt]
      It has the following neighbors at hop 2 (Indirectly connected by shared neighbors):\\[3pt]
      N1: \quad \textcolor{blue}{\textless 2-hop neighbor 1 text feature\textgreater}, \textcolor{blue}{\textless 2-hop neighbor 1 image feature\textgreater}\\[3pt]
      N2: \quad \textcolor{blue}{\textless 2-hop neighbor 2 text feature\textgreater}, \textcolor{blue}{\textless 2-hop neighbor 2 image feature\textgreater}\\[3pt]
      ......\\[5pt]
      Based on the information provided, please determine whether a link exists between the two nodes. Answer \textcolor{blue}{"yes"} if a link exists or \textcolor{blue}{"no"} if it does not.
    \end{minipage} \\
    \bottomrule
  \end{tabular}
\end{table*}

\begin{figure*}[t]
\centering
\caption{A case from the Movies dataset in the NC task that Mario identifies as preferring Text+Image modality information.}
\label{fig:case1}
\begin{casebox}{Case Study: Node Classification-Text+Image-Movies}

\textbf{Anchor node raw text:}\\[2pt]
In a strange dark age based on Celtic myths, the Divine Empire's path of conquest seems unstoppable... until a savage priest makes a critical mistake while attempting to resurrect a Demon Lord! Now the scales of fate tip in the other direction.

\medskip

\textbf{Label list:}\\[2pt]

\begin{tabular}{@{}ll@{}}
  'A\&E Home Video'                       & 'Art House \& International' \\
  'BBC'                                   & 'Blu-ray' \\
  'Boxed Sets'                            & 'Classics' \\
  'Criterion Collection'                  & 'Fully Loaded DVDs' \\
  'Genre for Featured Categories'         & 'HBO' \\
  'Holidays \& Seasonal'                  & 'Independently Distributed' \\
  'Movies'                                & 'Music Artists' \\
  'Musicals \& Performing Arts'           & 'Paramount Home Entertainment' \\
  'Science Fiction \& Fantasy'            & 'Studio Specials' \\
  'TV'                                    & 'Walt Disney Studios Home Entertainment'
\end{tabular}

\medskip

\textbf{ChatGPT-5.1-Thinking:}\\[2pt]
\predwrong{Science Fiction \& Fantasy\ \textbf{\xmark}}

\medskip

\textbf{Gemini-3-Pro:}\\[2pt]
\predwrong{Science Fiction \& Fantasy\ \textbf{\xmark}}

\medskip

\textbf{Qwen3-Max:}\\[2pt]
\predwrong{Science Fiction \& Fantasy\ \textbf{\xmark}}

\medskip

\textbf{Mario-8B:}\\[2pt]
\predcorrect{Genre for Featured Categories\ \textbf{\cmark}}

\end{casebox}
\end{figure*}

\begin{figure*}[h] 
    \centering
    \includegraphics[width=0.2\linewidth]{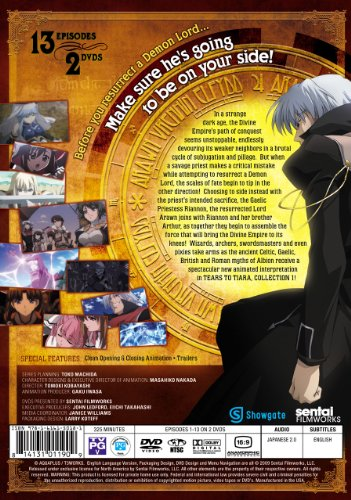} 
    \caption{Anchor node's image in Figure~\ref{fig:case1}.}
    \label{fig:case1figure}
\end{figure*}

\begin{figure*}[t]
\centering
\caption{A case from the Movies dataset in the NC task that Mario identifies as preferring Text-only modality information.}
\label{fig:case2}
\begin{casebox}{Case Study: Node Classification-Text-only-Movies}

\textbf{Anchor node raw text:}\\[2pt]
Roots of Human Behavior; Description: While human history is usually studied from the perspective of a few hundred years, anthropologists consider deeper causes for the ways we act. Now, in these 12 engrossing lectures, you'll join an expert anthropologist as she opens an enormous window of understanding for you into the thrilling legacy left by our primate past.

\medskip

\textbf{Label list:}\\[2pt]

\begin{tabular}{@{}ll@{}}
  'A\&E Home Video'                       & 'Art House \& International' \\
  'BBC'                                   & 'Blu-ray' \\
  'Boxed Sets'                            & 'Classics' \\
  'Criterion Collection'                  & 'Fully Loaded DVDs' \\
  'Genre for Featured Categories'         & 'HBO' \\
  'Holidays \& Seasonal'                  & 'Independently Distributed' \\
  'Movies'                                & 'Music Artists' \\
  'Musicals \& Performing Arts'           & 'Paramount Home Entertainment' \\
  'Science Fiction \& Fantasy'            & 'Studio Specials' \\
  'TV'                                    & 'Walt Disney Studios Home Entertainment'
\end{tabular}

\medskip

\textbf{ChatGPT-5.1-Thinking:}\\[2pt]
\predwrong{TV\ \textbf{\xmark}}

\medskip

\textbf{Gemini-3-Pro:}\\[2pt]
\predcorrect{Genre for Featured Categories\ \textbf{\cmark}}

\medskip

\textbf{Qwen3-Max:}\\[2pt]
\predwrong{Movies\ \textbf{\xmark}}

\medskip

\textbf{Mario-8B:}\\[2pt]
\predcorrect{Genre for Featured Categories\ \textbf{\cmark}}

\end{casebox}
\end{figure*}

\begin{figure*}[h] 
    \centering
    \includegraphics[width=0.3\linewidth]{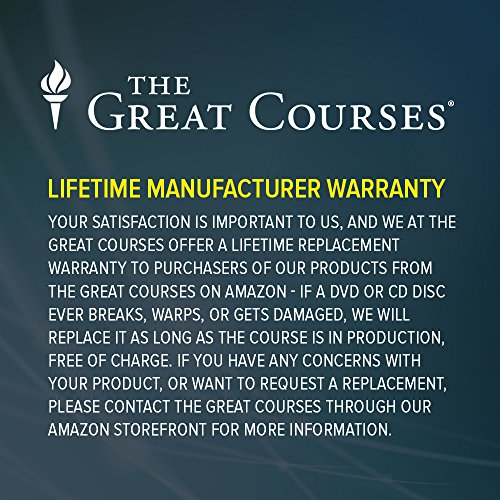} 
    \caption{ Anchor node’s image in Figure~\ref{fig:case2}}
    \label{fig:case2image}
\end{figure*}

\begin{figure*}[t]
\centering
\caption{A case from the Movies dataset in the NC task that Mario identifies as preferring Image-only modality information.}
\label{fig:case3}
\begin{casebox}{Case Study: Node Classification-Image-only-Movies}

\textbf{Anchor node raw text:}\\[2pt]
Castle: Season 6; Description: After Castle's stunning romantic proposal to Beckett, what happens next? TV's magnetic crime-fighting couple faces a whole new set of challenges as they juggle wedding plans and their most intriguing cases yet in ABC's CASTLE: THE COMPLETE SIXTH SEASON. Beckett's new job with the Justice Department takes her away from the wisecracking love of her life. But Castle's devotion to his new fiancee -- and her fascinating line of work -- jeopardizes her career and creates a chain of events that might separate them forever. Back on the home front, Castle is none too pleased to discover his daughter has seemingly been captivated by, and now living with, her new, free-spirited boyfriend.

\medskip

\textbf{Label list:}\\[2pt]

\begin{tabular}{@{}ll@{}}
  'A\&E Home Video'                       & 'Art House \& International' \\
  'BBC'                                   & 'Blu-ray' \\
  'Boxed Sets'                            & 'Classics' \\
  'Criterion Collection'                  & 'Fully Loaded DVDs' \\
  'Genre for Featured Categories'         & 'HBO' \\
  'Holidays \& Seasonal'                  & 'Independently Distributed' \\
  'Movies'                                & 'Music Artists' \\
  'Musicals \& Performing Arts'           & 'Paramount Home Entertainment' \\
  'Science Fiction \& Fantasy'            & 'Studio Specials' \\
  'TV'                                    & 'Walt Disney Studios Home Entertainment'
\end{tabular}

\medskip

\textbf{ChatGPT-5.1-Thinking:}\\[2pt]
\predwrong{TV\ \textbf{\xmark}}

\medskip

\textbf{Gemini-3-Pro:}\\[2pt]
\predwrong{TV\ \textbf{\xmark}}

\medskip

\textbf{Qwen3-Max:}\\[2pt]
\predwrong{TV\ \textbf{\xmark}}

\medskip

\textbf{Mario-8B:}\\[2pt]
\predcorrect{Boxed Sets\ \textbf{\cmark}}

\end{casebox}
\end{figure*}

\begin{figure*}[h] 
    \centering
    \includegraphics[width=0.2\linewidth]{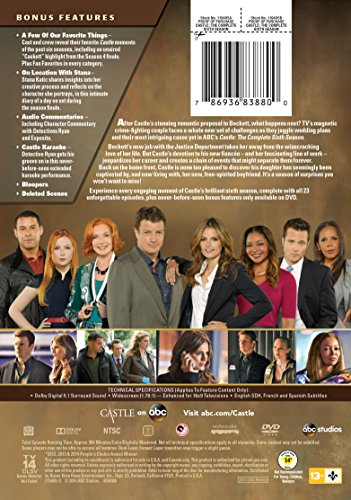} 
    \caption{Anchor node’s image in Figure~\ref{fig:case3}}
    \label{fig:case3image}
\end{figure*}

\begin{figure*}[t]
\centering
\caption{A case from the Toys dataset in the LP task that Mario identifies as preferring Text+Image modality information.}
\label{fig:case4}
\begin{caseboxpurple}{Case Study: Link Prediction-Text+Image-Toys}

\textbf{Node pair raw text:}\\[2pt]

\textit{Node 1:}\\[2pt]
The Crazy Scientist series, a collection of science tricks, was created by a joint venture of 2 crazy scientists and the Purple Cow. A combination bound to create an excellent and yet crazy experience! The Crazy Scientist Young Researches is a set of science tricks for kids to try out and discover 20 fun and fascinating facts about the world around you. Create excellent and crazy experiences that can be enjoyed by the entire family! Each science trick comes with a simple yet clever scientific explanation. A perfect STEAM gift! Challenge your brainpower and make intriguing discoveries about the world around us. Have fun experimenting and learning with the Young Researches amazing activities. Provide children the opportunity to become real researchers and follow easy instructions of science experiments that can be conducted using common household materials. Whats included? The box contains 20 activity cards with detailed instructions. Recommended for children ages 6 and up. Some science tricks may require adult supervision as indicated.\\[6pt]

\textit{Node 2:}\\[2pt]
Learn the scientific principles behind optical illusions with the 4M Illusion Science kit. Experiment with 20 classic optical illusions included in this kit. The kit includes illusion trick cards, spinning top with illusion cards, 3D picture cards, markers, 3D glasses, and more. A 20-page instruction book is included, describing the science of optical illusions and how to create a wide range of illusory effects. Perfect for young scientists with an interest in optics. Recommended for ages 7 years and up.

\medskip\medskip

\textbf{Label list:}\\[2pt]
\begin{tabular}{@{}ll@{}}
  \texttt{'yes'} & The two toys are co-purchased. \\
  \texttt{'no'}  & The two toys are not co-purchased.
\end{tabular}

\medskip

\textbf{ChatGPT-5.1-Thinking:}\\[2pt]
\predwrong{No\ \textbf{\xmark}}

\medskip

\textbf{Gemini-3-Pro:}\\[2pt]
\predcorrect{Yes\ \textbf{\cmark}}

\medskip

\textbf{Qwen3-Max:}\\[2pt]
\predcorrect{Yes\ \textbf{\cmark}}

\medskip

\textbf{Mario-8B:}\\[2pt]
\predcorrect{Yes\ \textbf{\cmark}}

\end{caseboxpurple}
\end{figure*}

\begin{figure*}[h]
    \centering
    \begin{subfigure}[b]{0.1\linewidth}
        \centering
        \includegraphics[width=\linewidth]{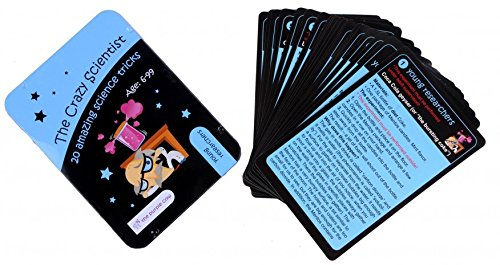}
        \caption{Node 1.}
        \label{fig:img-left}
    \end{subfigure}
    \begin{subfigure}[b]{0.1\linewidth}
        \centering
        \includegraphics[width=\linewidth]{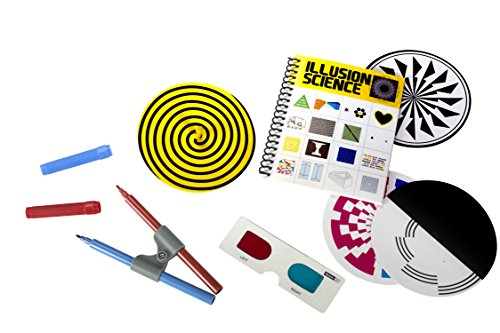}
        \caption{Node 2.}
        \label{fig:img-right}
    \end{subfigure}
    \caption{Node Pair's images in Figure~\ref{fig:case4}.}
    \label{fig:case4image}
\end{figure*}

\begin{figure*}[t]
\centering
\caption{A case from the CDs dataset in the LP task that Mario identifies as preferring Text-only modality information.}
\label{fig:case5}
\begin{caseboxpurple}{Case Study: Link Prediction-Text-only-CDs}

\textbf{Node pair raw text:}\\[2pt]

\textit{Node 1:}\\[2pt]
You Can Do It Yoga for MS Volume 2 DVD; Description: This DVD contains 2 complete classes. The first is a beginner/gentle yoga class. It includes some floor poses and some standing poses along with a guided meditation. Runtime: 54 minutes The second class is a beginner/intermediate yoga class. It includes some floor poses and some standing and balancing poses along with a guided meditation. Runtime: 50 minutes,This DVD contains 2 complete classes.\\[6pt]

\textit{Node 2:}\\[2pt]
Thoughts Become Things; Description: You create your own reality and by changing your thoughts, words, and actions in the simplest of ways, you can create fantastic change. - Mike Dooley".

\medskip\medskip

\textbf{Label list:}\\[2pt]
\begin{tabular}{@{}ll@{}}
  \texttt{'yes'} & The two CDs are co-purchased. \\
  \texttt{'no'}  & The two CDs are not co-purchased.
\end{tabular}

\medskip

\textbf{ChatGPT-5.1-Thinking:}\\[2pt]
\predwrong{No\ \textbf{\xmark}}

\medskip

\textbf{Gemini-3-Pro:}\\[2pt]
\predcorrect{Yes\ \textbf{\cmark}}

\medskip

\textbf{Qwen3-Max:}\\[2pt]
\predwrong{No\ \textbf{\xmark}}

\medskip

\textbf{Mario-8B:}\\[2pt]
\predcorrect{Yes\ \textbf{\cmark}}

\end{caseboxpurple}
\end{figure*}

\begin{figure*}[h]
    \centering
    \begin{subfigure}[b]{0.2\linewidth}
        \centering
        \includegraphics[width=\linewidth]{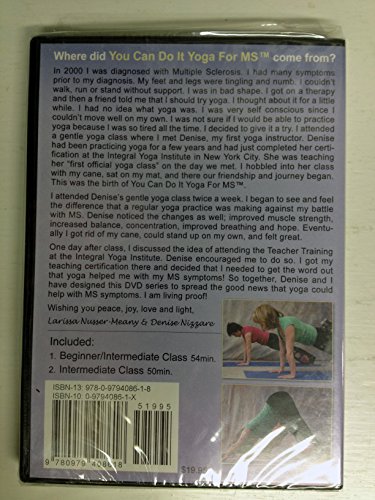}
        \caption{Node 1.}
        \label{fig:img-left}
    \end{subfigure}
    \begin{subfigure}[b]{0.2\linewidth}
        \centering
        \includegraphics[width=\linewidth]{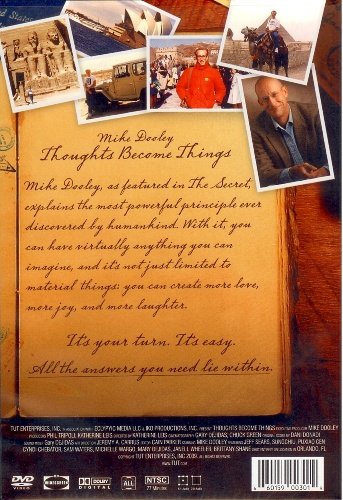}
        \caption{Node 2.}
        \label{fig:img-right}
    \end{subfigure}
    \caption{Node Pair’s images in Figure~\ref{fig:case5}.}
    \label{fig:case5image}
\end{figure*}

\begin{figure*}[t]
\centering
\caption{A case from the CDs dataset in the LP task that Mario identifies as preferring Image-only modality information.}
\label{fig:case6}
\begin{caseboxpurple}{Case Study: Link Prediction-Image-only-CDs}

\textbf{Node pair raw text:}\\[2pt]

\textit{Node 1:}\\[2pt]
Howard Lovecraft And The Frozen Kingdom; Description: After visiting his father in Arkham Sanitarium, young Howard Lovecraft ignores his father\&\#146s ominous warning and uses the legendary Necronomicon to open a portal to a strange, frozen world filled with horrifying creatures and grave danger. Alone and scared, Howard befriends a hideous creature he names Spot who becomes his companion throughout their treacherous journey across the Frozen Kingdom.\\[6pt]

\textit{Node 2:}\\[2pt]
A Serbian Film (Uncut) by Srdjan Todorovic; Description: Milos, a retired adult film star, leads a normal family life with his wife Maria and six-year old son Petar in tumultuous Serbia, trying to make ends meet. A sudden call from his former colleague Layla will change everything. Aware of his financial problems, Layla introduces Milos to Vukmir - a mysterious, menacing and politically powerful figure in the adult film business. A leading role in Vukmir's production will provide financial support to Milos and his family for the rest of their lives.

\medskip\medskip

\textbf{Label list:}\\[2pt]
\begin{tabular}{@{}ll@{}}
  \texttt{'yes'} & The two CDs are co-purchased. \\
  \texttt{'no'}  & The two CDs are not co-purchased.
\end{tabular}

\medskip

\textbf{ChatGPT-5.1-Thinking:}\\[2pt]
\predcorrect{No\ \textbf{\cmark}}

\medskip

\textbf{Gemini-3-Pro:}\\[2pt]
\predcorrect{No\ \textbf{\cmark}}

\medskip

\textbf{Qwen3-Max:}\\[2pt]
\predwrong{Yes\ \textbf{\xmark}}

\medskip

\textbf{Mario-8B:}\\[2pt]
\predcorrect{No\ \textbf{\cmark}}

\end{caseboxpurple}
\end{figure*}

\begin{figure*}[h]
    \centering
    \begin{subfigure}[b]{0.15\linewidth}
        \centering
        \includegraphics[width=\linewidth]{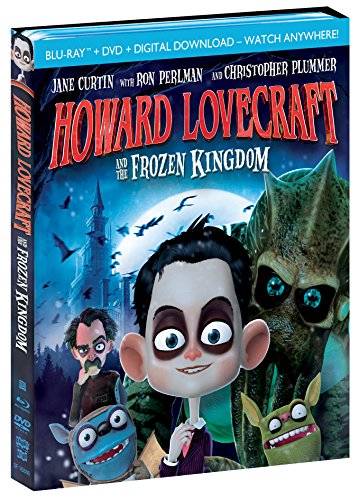}
        \caption{Node 1.}
        \label{fig:img-left}
    \end{subfigure}
    \begin{subfigure}[b]{0.15\linewidth}
        \centering
        \includegraphics[width=\linewidth]{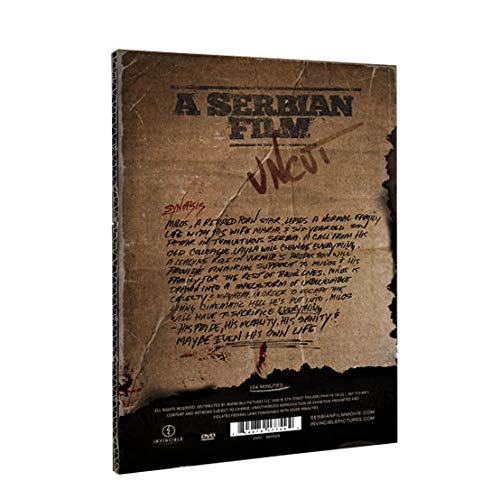}
        \caption{Node 2.}
        \label{fig:img-right}
    \end{subfigure}
    \caption{Node Pair’s images in Figure~\ref{fig:case6}}
    \label{fig:case6image}
\end{figure*}

\subsection{Case Study} 
To better understand how Mario behaves on multimodal graphs, we conduct a qualitative case study on both tasks. We compare Mario-8B against three strong closed-/API-based L(V)LMs—ChatGPT-5.1-Thinking, Gemini-3-Pro, and Qwen3-Max—on several representative nodes and node pairs drawn from the Movies, Toys, and CDs graphs (Figs.~\ref{fig:case1}–\ref{fig:case6image}). 
Because these models are accessed only through high-level APIs, we cannot inject special feature tokens when prompting as we do for Mario. Instead, we adopt a uniform and conservative prompting protocol: for each case, we present the anchor node (or node pair) together with its neighbors using the same high-level templates as in Table~\ref{tab:prompt_templates}, and we input each neighbor’s raw text and image jointly to ensure a fair comparison. 

For the node classification case shown in Figs.~\ref{fig:case2}–\ref{fig:case2image}, the anchor node’s text describes the content of a lecture series, whereas the associated image focuses almost entirely on after-sales information (lifetime warranty and replacement policy) and provides very little semantic signal about the lecture itself. Mario’s MAPR, conditioned on both the anchor’s multimodal features and its local subgraph, correctly infers that the image is not the preferred modality for this classification task and routes the node through a text-centric template. This decision matches the underlying graph semantics and illustrates that the router is able to down-weight visually salient but task-irrelevant information.

Across all the illustrated cases, Mario’s behavior is consistently competitive with, and sometimes superior to, strong closed-source L(V)LMs. In several examples, all closed models converge to the same intuitive but graph-inconsistent label, while Mario is the only method that predicts the correct class—for instance, in the case of Figs.~\ref{fig:case1}–\ref{fig:case1figure}, where Mario is the only model that assigns the ground-truth category and all other systems fail. These qualitative results further corroborate that Mario is an effective and reliable framework for multimodal graph reasoning.

\end{document}